\documentclass[journal, twoside, twocolumn]{IEEEtran}
\ifCLASSOPTIONcompsoc
  \usepackage[nocompress]{cite}

\else
  \usepackage{cite}
\fi
\ifCLASSINFOpdf
\else
\fi
\usepackage{amsmath}
\usepackage{tikz}
\usepackage{array}
\usepackage{times}
\usepackage[caption = false]{subfig}
\usepackage{helvet}
\usepackage{subfig}
\usepackage{amssymb}
\usepackage{amsmath}
\usepackage{amssymb}
\usepackage{courier}
\usepackage{multirow}
\usepackage{mathrsfs}
\usepackage{graphicx}
\usepackage{enumitem}
\usepackage{graphicx}
\usepackage{blindtext}
\usepackage{algorithm}
\usepackage{algorithmic}
\usepackage{lineno,hyperref}
\usepackage[marginal]{footmisc}

\usepackage[utf8]{inputenc}
\usepackage[english]{babel}
\usepackage{epstopdf}
\usepackage{array}
\usepackage{multirow}
\usepackage{booktabs}

\newtheorem{theorem}{Theorem}

\newtheorem{lemma}[theorem]{Lemma}
\newtheorem{proof}{Proof}
\newtheorem{remark}{Remark}
\begin{document}

\title{Unpaired Multi-View Graph Clustering with Cross-View Structure Matching}

\author{Yi~Wen,~Siwei~Wang,~Qing Liao,~Weixuan~Liang,~Ke~Liang,\\
 ~Xinhang~Wan,~Xinwang~Liu$^{\dagger}$,~\IEEEmembership{Senior~Member,~IEEE}
\IEEEcompsocitemizethanks{
\IEEEcompsocthanksitem Y. Wen, S. Wang, W. Liang, K. Liang, X. Wan and X. Liu are with School of Computer, National University of Defense Technology, Changsha, 410073, China. (E-mail: \{wenyi21,\, wangsiwei13,\,weixuanliang,\,  liangke22,\, wanxinhang,\, xinwangliu,\} @nudt.edu.cn).
\\
Q. Liao is with Department of Computer Science and Technology, Harbin Institute of Technology, Shenzhen, China (E-mail: liaoqing@hit.edu.cn).\\
\IEEEcompsocthanksitem $^{\dagger}$: Corresponding author.
\IEEEcompsocthanksitem Manuscript received Feb. 21, 2022.
}
}

\markboth{IEEE Transactions on Neural Networks and Learning Systems}%
{WEN \MakeLowercase{\textit{et al.}}: Unpaired Multi-View Graph Clustering with Cross-View Structure Matching}

\maketitle

\begin{abstract}
Multi-view clustering (MVC), which effectively fuses information from multiple views for better performance, has received increasing attention.
Most existing MVC methods assume that multi-view data are fully paired, which means that the mappings of all corresponding samples between views are pre-defined or given in advance. However, the data correspondence is often incomplete in real-world applications due to data corruption or sensor differences, referred as the data-unpaired problem (DUP) in multi-view literature. Although several attempts have been made to address the DUP issue, they suffer from the following drawbacks: 1) Most methods focus on the feature representation while ignoring the structural information of multi-view data, which is essential for clustering tasks; 2) Existing methods for partially unpaired problems rely on pre-given cross-view alignment information, resulting in their inability to handle fully unpaired problems; 3) Their inevitable parameters degrade the efficiency and applicability of the models. To tackle these issues, we propose a novel parameter-free graph clustering framework termed Unpaired Multi-view Graph Clustering framework with Cross-View Structure Matching (UPMGC-SM). Specifically, unlike the existing methods, UPMGC-SM effectively utilizes the structural information from each view to refine cross-view correspondences. Besides, our UPMGC-SM is a unified framework for both the fully and partially unpaired multi-view graph clustering. Moreover, existing graph clustering methods can adopt our UPMGC-SM to enhance their ability for unpaired scenarios.
Extensive experiments demonstrate the effectiveness and generalization of our proposed framework for both paired and unpaired datasets.

\end{abstract}

\begin{IEEEkeywords}
Multi-view Clustering, graph learning, unpaired data, graph fusion.
\end{IEEEkeywords}

\section{Introduction}\label{sec:introduction}

%
%
%
%

\IEEEPARstart{W}ITH the rapid development of information technology, data are often generated from different sources in real-world scenarios \cite{liang2022reasoning,liang2023knowledge,liang2021mka,liang2023abslearn}. For instance, the same news can be described from different views, \textit{i.e.,} textual reports and visual pictures \cite{ICRN,xihong_tnnls,AKGR,liuyue_DCRN}. Consequently, how to effectively utilize information from different views is the key to enhancing the final clustering performance. Multi-view clustering (MVC), one of the most classical unsupervised tasks \cite{huang2019auto,wang2015robust,zhang2018generalized,guo2017distributed, ABSLearn}, has drawn increasing attention by effectively leveraging the complementary and consensus information from multiple views.   

Graph structures, which can well describe the relationships of pairwise data, are widely adopted in the field of MVC \cite{chen2022neuroadaptive,nie2011spectral,nie2020subspace,li2020multi,nie2016constrained,zhang2020consensus}. In general, graph-based multi-view clustering achieves remarkable performance with two main procedures \cite{ ren2020simultaneous,wang2022align}, \textit{i.e.,} generating a graph for each view and operating graph fusion on the individual graph. 
For example, \cite{nie2016parameter} generates the optimal consensus graph matrix from the linear combination of the base graph matrices from multiple views.

\begin{figure*}[t] 
\centering 
\includegraphics[width=1\textwidth]{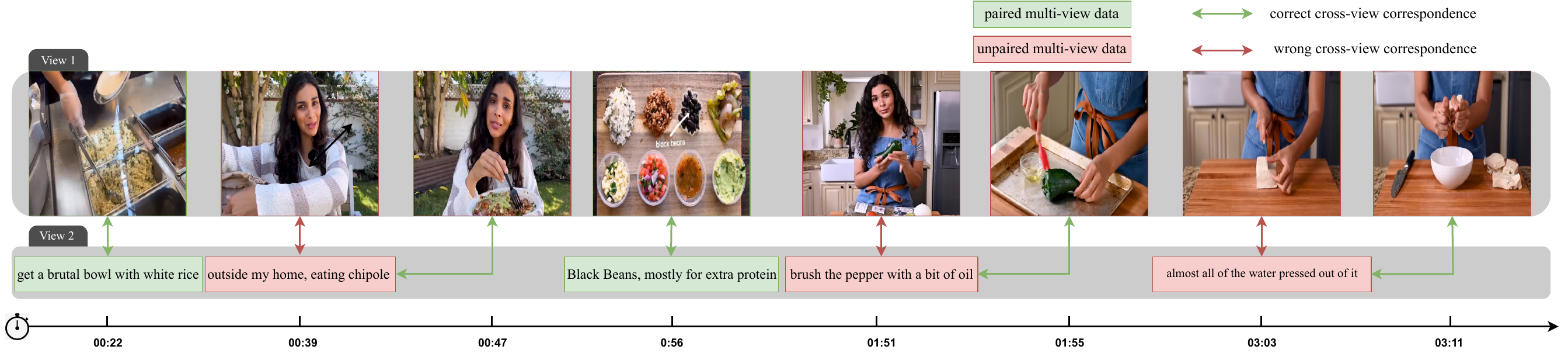} 
\caption{A typical multi-view data-unpaired problem (DUP) appears in a raw instructional video. In this case, the visible video and narrator text represent different views separately. The DUP is caused by the fact that the narrator's description is not synchronized with the visual content. The green and red objects indicate the paired and unpaired multi-view data. Example from \url{https://www.youtube.com/watch?v=UBkAOSlJz4g}} 
\label{Fig.example} 
\end{figure*}

Although the above MVC methods have achieved promising performance \cite{zhang2018binary,li2019flexible,li2017liquid,liu2022fast,zhang2021joint}, they often fail when multi-view data is unpaired because incorrect alignment information between views can mislead the final cluster partition \cite{lin2022tensor}. Specifically, the unpaired data implies, for a given multi-view dataset denotes the data of the $v$-th view, the data $x_i^v, x_i^u$ with the same index in different views can represent different objects, termed \textbf{D}ata-\textbf{U}npaired \textbf{P}roblem (DUP). In a real-world scenario, such a situation could commonly happen because different views often collect from various sources separately, resulting in a lack of additional alignment information between views. A typical example of DUP is an instructional video in Figure \ref{Fig.example}, where the visible video and narrator text represent different views. In this case, the narrator's description is out of synchrony with the visual content, causing the data-unpaired problem.

Several approaches have been proposed to solve this problem,
\cite{huang2020partially} uses a differentiable surrogate of the Hungarian algorithm to learn the correspondence of the unaligned data in a latent space learned by a neural network. \cite{yang2021partially} proposes a noise-robust contrastive loss to establish the view correspondence. Based on the framework of non-negative matrix factorization, \cite{yu2021novel} utilizes both feature and local structural constraints to establish the cross-view correspondence. 

Although remarkable success has been made, the above methods suffer from the following drawbacks: Firstly, most methods are devoted to learning the feature representation, which ignores the importance of structural information for the clustering task; Secondly, there is a lack of a unified framework to address both the above problem, which leaves us space to explore; Thirdly, their inevitable hyper-parameter selection dramatically reduces the efficiency and applicability of the model because there is no prior knowledge to guide us on how to choose a proper reduction dimension.

To tackle these issues, we propose a parameter-free graph clustering framework termed Unpaired Multi-view Graph Clustering with Cross-View Structure Matching (UPMGC-SM). Unlike existing methods, UPMGC-SM adopts a flexible graph clustering algorithm to generate individual graphs on each view. Then, a shared space is established to solve the problem that cross-view data are not in the same metric space and cannot be metricized. Besides, UPMGC-SM captures the structural information from views to refine the mapping relationship by using the projected fixed-point method and doubly stochastic projection. Finally, the graph fusion is processed with the established mappings to obtain the common graph structure. Note that our UPMGC-SM is a unified framework for both fully and partially unpaired multi-view graph clustering. Moreover, the flexible framework can adapt to any existing graph clustering and get desirable results. We summarize the contributions of this paper as follows:
\begin{itemize}
\item We propose a new paradigm for graph-based MVC termed \textbf{D}ata-\textbf{U}npaired \textbf{P}roblem (DUP). Our solution to this problem is of great significance and can avoid the strong assumption of the alignment of cross-view data in data collection.
\item We propose a parameter-free clustering framework termed Unpaired Multi-view Graph Clustering with Cross-View Structure Matching (UPMGC-SM) to address the above issues, which can efficiently refine cross-view correspondence and get the fused graph simultaneously. To the best of our knowledge, this is the first solution that can simultaneously solve the partially and fully unpaired problem in the field of MVC.
\item  Extensive experiments on nine datasets demonstrate the effectiveness and generalization of our proposed framework. Besides, we theoretically prove the convergence of our algorithm.
\end{itemize}

\vspace{0.3cm}

\section{Related Work}
Based on the given alignment information between views,  multi-view clustering (MVC) can be divided into two categories: MVC on fully paired data and MVC on unpaired data. At the same time, unpaired data can be further divided into partially unpaired data and fully unpaired data, as shown in Figure \ref{Fig.main}.

\subsection{Multi-View Clustering on Fully Paired Data}
\begin{figure}[t]
\begin{center}{
    \centering
     \subfloat[\scriptsize{Fully Paired}\label{fulp}]{\includegraphics[width=0.152\textwidth]{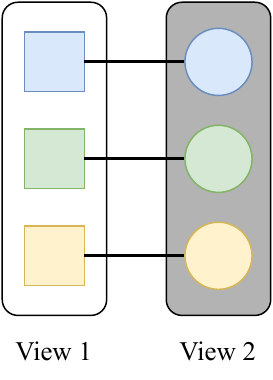}}
    \subfloat[\scriptsize{Partially Unpaired}\label{parup}]{\includegraphics[width=0.152\textwidth]{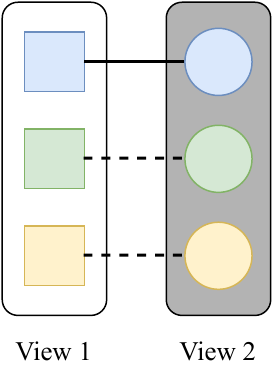}}
    \subfloat[\scriptsize{Fully Unpaired}\label{fulup}]{\includegraphics[width=0.152\textwidth]{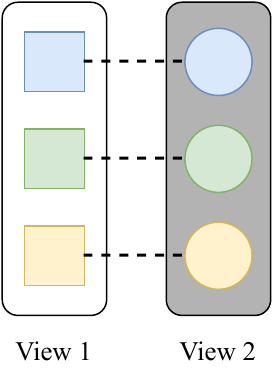}}
    \caption{Three types of cross-view data. The connection indicates the correspondence relationship between cross-view data. The solid line in the figure indicates the cross-view mapping we know, while the dashed line indicates the unknown cross-view mapping.}
    \label{Fig.main}}
\end{center}
\end{figure}

Fully paired data implies that all mapping relationships are given for every two cross-view data, as shown in Figure \ref{fulp}. Multi-view clustering on fully paired data has been widely studied \cite{LiangTNNLS,2023arXiv230602389W,wan2023autoweighted,10.1145/3503161.3547864}, which can be broadly classified into five categories. (1) Multiple kernel clustering \cite{wang2020smoothness,gonen2008localized,gonen2011multiple ,Li2016Multiple,liu2020optimal ,liu2021hierarchical}. Multiple kernel clustering often predefines a set of base kernels on individual views and then optimally fuses the weights of these kernels to enhance the final clustering performance.
(2) Graph-based multi-view clustering \cite{nie2021learning,lin2021multi,liu2010large,ma2020towards,zhan2017graph}. Graph-based multi-view clustering aims to seek a fused graph from different views, and the final clustering results are obtained through a spectral decomposition process.
(3) Multi-view subspace clustering \cite{tang2018learning,zhang2021joint,adler2015linear,luo2018consistent,zhang2015low,peng2015unified,kang2020partition}. Subspace clustering algorithms assume that all views share a latent low-dimensional representation, and the final clustering result is obtained by learning the shared representation. (4) Multi-view  based on non-negative matrix factorization (NMF) \cite{shao2015multiple}. NMF-based clustering decomposes the data matrix from multiple sources into two non-negative components, i.e., consensus coefficient matrix and view-specific base matrix.  For instance, Liu et al.\cite{liu2013multi} propose a novel NMF-based multi-view clustering algorithm to obtain the consensus clustering partition in multiple views by factorization.
    Wang et al.\cite{wang2017exclusivity} propose a diverse NMF approach with a novel defined diversity term to reduce the redundancy among multi-view representations. 
    Nie et al.\cite{nie2020auto} introduce a new co-clustering method to decrease information loss in matrix factorization. 
    Further, the work in \cite{cai2020semi} studies a non-negative matrix factorization with orthogonality constraints, which aims to discover the intra-view diversity and a set of independent bases. (5) Multi-view ensemble clustering. Multi-view ensemble clustering seeks to fuse information on multiple views into a more robust and attractive clustering result.  For example, Tao et al.\cite{tao2017ensemble} leverage the complementary information of multi-view data by generating Basic Partitions (BPs) for each view individually and seeking a consensus partition among all the BPs.
    Moreover, Liu et al.\cite{liu2017spectral} propose a novel spectral ensemble clustering method to leverage the advantages of a co-association matrix in highly efficient information integration.

Although massive methods have been proposed in multi-view clustering, most of them are based on the assumption that the cross-view data are fully paired, which is often difficult to achieve in real applications.

\noindent

\subsection{Multi-View Clustering on Unpaired Data}
Unpaired MVC is known as a challenging task from two main aspects. Firstly, it is hard to learn the correct correspondence of the cross-view data without any prior information. Secondly, it is hard to directly calculate the distance of the cross-view data since the features from diverse sources are usually in different dimensions. 

\begin{figure}[t] 
\centering 
\includegraphics[width=0.49\textwidth]{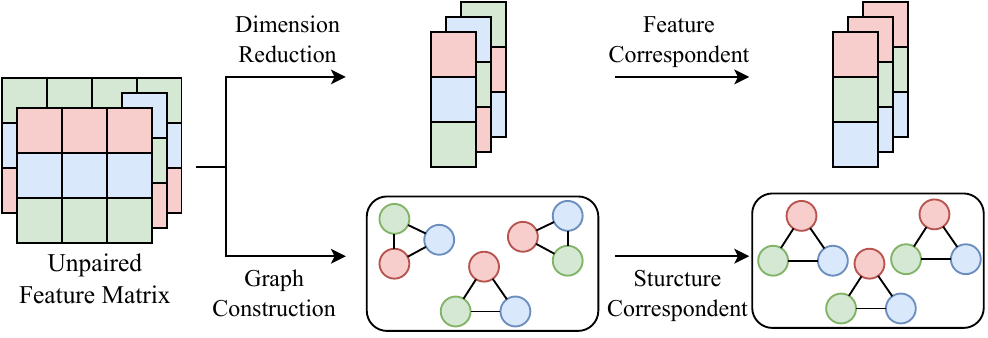} 
\caption{The main processes of the feature correspondence method (above) and structure correspondence method (below). The different matrices denote the feature matrix from different views. The object with the same color represents the same object from different views.} 
\label{Fig.fea} 
\end{figure}

To tackle these issues, dimensionality correspondence techniques, including feature correspondence and structure correspondence, came into existence. Both of the methods are based on the solution of finding a latent common space to calculate the discrepancy between cross-view data. As shown in Figure \ref{Fig.fea}, the strategy detail of the two correspondence techniques is different. Feature correspondence aims to find the correspondence by maximizing feature similarities between different views, while the critical idea of structure correspondence is to identify correspondences by the similarity of their features from different views.

\subsubsection{Partially Unpaired}
Partially unpaired data means that a portion of the multi-view data is unpaired, as is shown in Figure \ref{parup}. In fact, the partially unpaired problem can be considered a special case of the semi-supervised problem, where partially paired data can provide supervised view alignment information. Several methods for partially unpaired data have been proposed by adopting the feature correspondence method.  \cite{zhang2015constrained} construct different types of constraint matrices in optimization to learn a desirable correspondence. \cite{huang2020partially}  design a differentiable alignment module and integrate it into a deep neural network to reconstruct the view alignment relationship. \cite{yang2021partially} proposes multi-view contrastive learning with a noise robust loss learning model such that paired samples are in the representation space close to each other, while the unpaired samples are away from each other. 

Several methods for partially unpaired data have been proposed by adopting different prior information. One pioneer, Qian et al.\cite{qian2013multi}, first introduced the weak side information of cross-view must-links and cross-view cannot-links to apply to multi-view classification situations. Chen et al.\cite{chen2012unified} present a general dimensionality reduction framework for semi-paired and semi-supervised multi-view data, which naturally generalizes existing related works by using different kinds of prior information. Muti-view hashing aims to model the multi-view feature correlations and preserve them in binary hash codes. Representatively, Zheng et al.\cite{zheng2020adaptive} propose an unsupervised adaptive partial Multi-view Hashing method to handle the partial-view hashing problem for efficient social image retrieval. 

However, these methods focus on learning the feature representation of the data, which ignores the importance of structural information for the clustering task. Moreover, when the alignment rate is low, these methods usually suffer from poor performance because of the lack of sufficient alignment information to recover a better correspondence mapping.

\subsubsection{Fully Unpaired}
    Fully unpaired data implies none of the cross-view correspondence mappings is given, as is shown in Figure \ref{fulup}. Based on the non-negative matrix factorization, one pioneer work, \cite{yu2021novel} aims to solve the fully unpaired problem, which uses both the feature information and structural information to solve the problem. 
The final cross-view correspondence mapping can be calculated by solving the following problem:

\begin{equation}
    \begin{aligned}
    \min _{\mathbf{W}^a, \mathbf{H}^a, \mathbf{P}^{a b}} &\sum_{a=1}^v\left\|\mathbf{X}^a-\mathbf{W}^a\left(\mathbf{H}^a\right)^T\right\|_{\mathcal{F}}^2+2 \alpha_a \mathcal{R}^a \\
    &+\sum_{1 \leq a<b \leq v} \lambda_a\left\|\left(\mathbf{P}^{a b}\right)^T \mathbf{~S}^a\left(\mathbf{P}^{a b}\right)-\mathbf{S}^b\right\|_{\mathcal{F}}^2 \\
&+\sum_{1 \leq a<b \leq v} \mu_a\left\|\mathbf{P}^{a b} \mathbf{H}^b-\mathbf{H}^a\right\|_{\mathcal{F}}^2 \text{,}\\
&\text { s.t. } \forall 1 \leq a \leq v, \mathbf{~W}^a \geq 0, \\
&\mathbf{H}^a \geq 0 ,\forall 1 \leq a<b \leq v, \mathbf{P}^{a b} \geq 0,
\end{aligned}
\end{equation}
where $s_{i j}^a=\left\|\mathbf{x}_i^a-\mathbf{x}_j^a\right\|^2$, $\mathbf{P}^{ab}$ is the correspondence mapping matrix of the $a$-th view and $b$-th view, $\alpha_a, \lambda_a$, and $\mu_a$ are the balance parameter of local structure, non-negative constraints, and structural correspondence mapping loss, respectively.

The algorithm focuses on the learning of the cross-view alignment matrix and obtains the final clustering results by the representations obtained on a single view, ignoring the complementary information that different views have, which degrades its clustering performance on some datasets. Besides, the abundant parameters in this method give an additional time cost and make it difficult to seek the algorithm's optimal solution.

\begin{figure*}[t] 
\centering 
\includegraphics[width=1\textwidth]{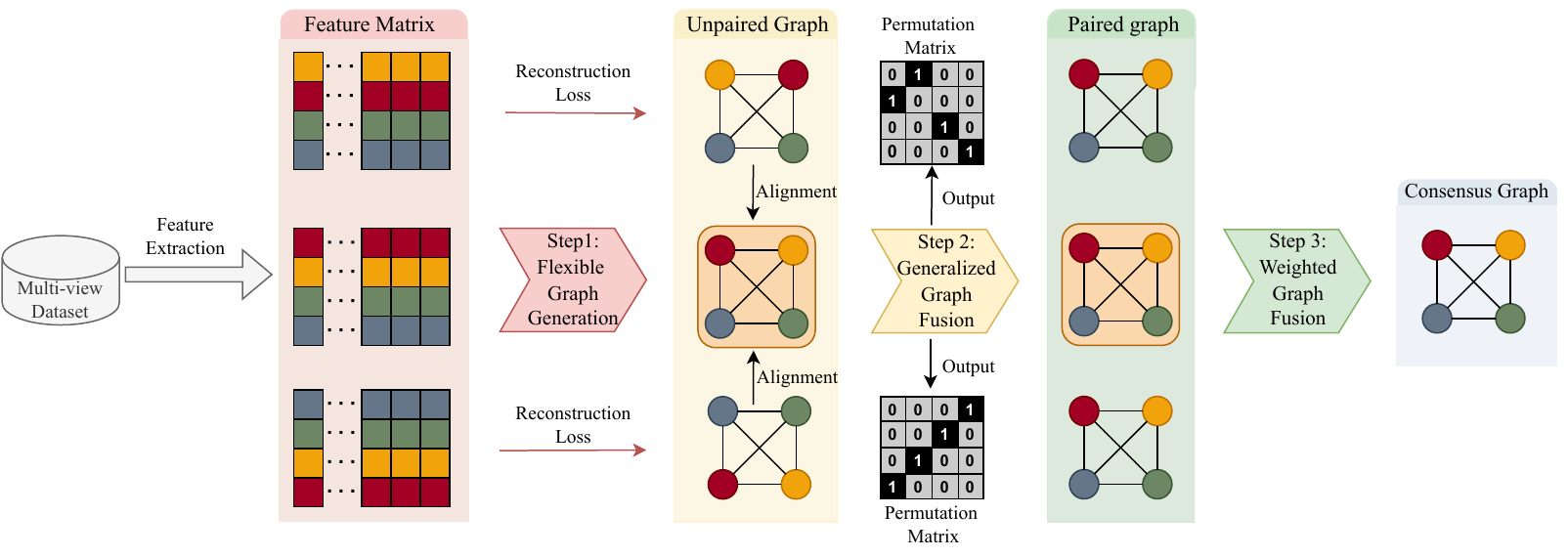} 
\caption{The framework of the proposed UPMGC-SM. As depicted in the figure, three main steps are adopted to obtain the final aligned graph. At the end of the algorithm, spectral clustering is implemented to obtain the final clustering results.} 
\label{Fig.main_algo} 
\end{figure*}

\section{Method}
In this section, we propose a parameter-free clustering framework termed Unpaired Multi-View Graph Clustering with Cross-View Structure Matching (UPMGC-SM). As shown in Figure \ref{Fig.main_algo}, three main steps are adopted to get the final cluster results. Besides, a brief summary of the UPMGC-SM is in the Implementation part. At the end of the section, we theoretically prove the convergence of our algorithm.

\subsection{Motivations}
As mentioned before, the main challenge for solving unpaired data problems is the different views on different metric spaces. Thus we can't directly measure the distance of the same sample of any two views. As a result, a question worth considering: \textbf{how to effectively find the correct correspondence of different views on different metric spaces?}

Several attempts have been proposed to solve the problem. However, most methods focus on the representation of data features, such as feature correspondences. Meanwhile, the structure of the data is rarely taken into account when learning representations, where \textbf{the omitted structural information is proven to be essential to the cluster task}, which has been well recognized by previous literature\cite{chen2019multi,wang2019spectral,huang2019ultra,wen2021global,liu2012robust,peng2018structured}. Moreover, feature correspondence will unavoidably encounter \textbf{additional hyper-parameter selection} because there is no prior knowledge to guide us on how to choose a proper reduction dimension. A common solution is to search the parameters over a large range for each dataset, which dramatically reduces the efficiency and applicability of the model.

Considering the weakness of the feature correspondence, we propose a novel framework based on the structure correspondence, which efficiently avoids the additional parameter choices. 

\subsection{Flexible Graph Generation}
     
 Given multi-view data matrices $\left\{\mathbf{X}_{i}\right\}_{i=1}^{v}$ with $n$ samples and $v$ views. As shown in Figure \ref{Fig.main}, the first step of the structural correspondence method is to construct individual graphs on each view by solving the following problem:
\begin{equation}
\label{fgg}
    \begin{aligned}
    \underset{\mathbf{S}_i, \mathbf{S}}{\rm{min}} \ & \mathcal{L}\left(\mathbf{X}_{i}, \mathbf{S}_i\right)+\Omega\left( \mathbf{S}_i, \mathbf{S}\right),\\
\text { s.t. } & \mathbf{S}_i \geq 0, \mathbf{S}_i^{\top} \mathbf{1}=\mathbf{1}, \mathbf{S} \geq 0, \mathbf{S}^{\top} \mathbf{1}=\mathbf{1},
\end{aligned}
\end{equation}
where $\mathbf{S}_i$ is the data self-representation matrix for the $i$-th view, $\mathcal{L}$ is the flexible reconstruction loss for learning the individual graph, and $\Omega$ is the process of the graph fusion. Considering the unpaired setting, in this step, we simply take the $\mathbf{X}_{i}$ of each view as input to get the view-independent graph. The next step is to refine the correspondence between different views. 

\subsection{Selection Principle for the Alignment View}
Before the graph matching process, the critical problem is to select the appropriate view for alignment. Colossal information loss will be caused when the most informative view has to align with the least informative one. Moreover, manually selecting the alignment view will bring an extra hyper-parameter. When the number of views is enormous, selecting an optimal hyper-parameter will bring a considerable time cost. As a result, choosing a reasonable evaluation index to assess the quality of the views is necessary. In this paper, We adopted the strategy of reconstruction loss minimization (RCM) to select the alignment view, which is calculated as  

\begin{equation} \label{cal loss}
\small
   \mathcal{L}_{con}^{i} = \mathcal{L}\left(\mathbf{X}_{i}, \mathbf{S}_i\right) \text{,}
\end{equation}
where $\mathcal{L}$ is the flexible reconstruction loss calculated by Eq. (\ref{fgg}). As a result, the view we select to align is 
\begin{equation}
\small
    i^* = \underset{i}{\arg\min} { \mathcal{L}_{con}^i}.
\end{equation}

\subsection{Generalized Graph Matching Framework}
In this subsection, we propose a generalized graph matching framework based on the structural information. For structural correspondence, we consider the following criteria: the internal structure of two graphs should be similar enough after a good mapping. As a result, the alignment of the graph structure can be achieved by optimizing the following formula:

\begin{equation}
\label{opt p 1}
\min _{\tilde{\mathbf{P}}_j}\left\|\tilde{\mathbf{S}}_{i^*}-\tilde{\mathbf{P}}^{\top}_j \tilde{\mathbf{S}}_{j} \tilde{\mathbf{P}}_j\right\|_{\mathbf{F}}^{2}, j \in \{1,\dots,v\}, j \ne i^{*},
\end{equation}

\noindent
\begin{remark}
Considering the fully unpaired and partial unpaired case, we only need to establish the cross-view correspondence between unpaired data rather than the whole data.  Stated differently, we only need to align the sub-blocks $\left\{\tilde{\mathbf{S}}_{i} \in \mathbb R ^{\tilde{n} \times \tilde{n}}\right\}_{i=1}^{v}$ of the original matrix  $\left\{\mathbf{S}_{i}\right\}_{i=1}^{v}$  ($\mathbf{S}_{i}$ contains the structural information among all data), where ${\tilde{n}}$ is the number of unpaired data and $\tilde{\mathbf{S}}_i$ corresponds to the structural information among the $\tilde{n}$ unpaired data of $i$-th view.
\end{remark}
where $\tilde{\mathbf{P}}_j$ is the corresponding permutation matrix of the $j$-th view, which satisfies 
$$
    \left\{\tilde{\mathbf{P}}_j \vert \tilde{\mathbf{P}}_j \mathbf{1} = \mathbf{1},\tilde{\mathbf{P}}_j^{\top} \mathbf{1}=\mathbf{1}, \tilde{\mathbf{P}}_j\in \left\{0, 1 \right\}^{n \times n}\right\} \text{.}
$$

The Eq.\eqref{opt p 1} can be easily transformed to 
\begin{equation}\label{opt P 2}
\begin{aligned}
\max _{\tilde{\mathbf{P}}_j}&\operatorname{Tr}\left(\tilde{\mathbf{S}}_{i^*}^{\top} \tilde{\mathbf{P}}_j^{\top} \tilde{\mathbf{S}}_{j} \tilde{\mathbf{P}}_j\right), \\
\text { s.t. } & \tilde{\mathbf{P}}_j \mathbf{1} = \mathbf{1}, \tilde{\mathbf{P}}_j^{\top} \mathbf{1}=\mathbf{1}, \tilde{\mathbf{P}}_j \in \left\{0, 1 \right\}^{n \times n}.
\end{aligned}
\end{equation}

 The inner graph structures can reach a maximum agreement by maximizing Eq. \eqref{opt P 2}.
 
 The problem in Eq. \eqref{opt P 2} is a classical NP-hard problem, and some methods \cite{mckay1981practical,ullmann1976algorithm} are proposed to find an exact solution. However, their time complexity is exponential in the worst case, which is unacceptable for us. Therefore, using a reasonable approximation method is much necessary. A common relaxation constraint is the space of partial permutation matrices
$$
    \boldsymbol{\Delta} = \left\{\tilde{\mathbf{P}}_j \ \vert \ \tilde{\mathbf{P}}_j \mathbf{1} = \mathbf{1},\tilde{\mathbf{P}}_j^{\top} \mathbf{1}=\mathbf{1}, \tilde{\mathbf{P}}_j \in \left[0, 1 \right]^{n \times n}\right\} \text{.}
$$

Then Eq. \eqref{opt P 2} can be transformed into

\begin{equation}\label{opt P 3}
\begin{aligned}
\small
\max _{\tilde{\mathbf{P}}_j}&\operatorname{Tr}\left(\tilde{\mathbf{S}}_{i^*}^{\top} \tilde{\mathbf{P}}_j^{\top} \tilde{\mathbf{S}}_{j} \tilde{\mathbf{P}}_j\right),  \\
\text { s.t. } & \tilde{\mathbf{P}}_j \mathbf{1} = \mathbf{1}, \tilde{\mathbf{P}}_j^{\top} \mathbf{1}=\mathbf{1}, \tilde{\mathbf{P}}_j \in \left[0, 1 \right]^{n \times n}.
\end{aligned}
\end{equation}

Eq. \eqref{opt P 3} is termed multi-view graph correspondence framework. To efficiently solve Eq. \eqref{opt P 3}, we adopt the Projected Fixed-Point Algorithm \cite{lu2016fast} to update $\tilde{\mathbf{P}}_j$ as follows,

\begin{equation}
\begin{aligned}
\label{opt P}
\tilde{\mathbf{P}}_j^{(t+1)}&=\frac{1}{2} \left(\tilde{\mathbf{P}}_j^{(t)}+ \boldsymbol{\Gamma}\left(\nabla f\left(\tilde{\mathbf{P}}_j^{(t)}\right)\right)\right)\\
&=\frac{1}{2} \left(\tilde{\mathbf{P}}_j^{(t)}+ \boldsymbol{\Gamma}\left( \tilde{\mathbf{S}}_{j} \tilde{\mathbf{P}}_j^{(t)} \tilde{\mathbf{S}}_{i^*}^{\top}\right)\right).
\end{aligned}
\end{equation}
where $t$ denotes the number of iterations and $\boldsymbol{\Gamma}$ denotes the double stochastic projection operator for given gradient matrix $\mathbf{G}=\nabla f\left(\tilde{\mathbf{P}}_j^{(t)}\right)$, and can be obtained by following two steps,
$$
\boldsymbol{\Gamma}_{1}(\mathbf{G})=\arg \min _{\mathbf{G}_{1}}\left\|\mathbf{G}-\mathbf{G}_{1}\right\|_{\mathbf{F}} \text {, s.t. } \mathbf{G}_{1} \mathbf{1}=\mathbf{1}, \mathbf{G}_{1}^{\top} \mathbf{1}=\mathbf{1} ,
$$
$$
\boldsymbol{\Gamma}_{2}(\mathbf{G})=\arg \min _{\mathbf{G}_2}\left\|\mathbf{G}-\mathbf{G}_{2}\right\|_{\mathbf{F}}, \quad \text { s.t. } \mathbf{G}_{2} \geq 0 .
$$
Both of the two projections have closed-formed solutions \cite{lu2016fast}, and the von Neumann successive projection lemma \cite{von1951functional} guarantees the convergence of the successive projection. After solving Eq. \eqref{opt P 3} to obtain the graph matrix, we apply the Sinkhorn operator to obtain the binary permutation matrix.


%

Meanwhile, we can use the identity $\mathbf{I} \in \mathbb R ^{ \left(n-\tilde{n}\right) \times \left(n-\tilde{n}\right)}$ to describe the correspondence of the paired data (since the known correspondence information can be considered as a one-to-one mapping), where $n$ and ${\tilde{n}}$ is the number of the entire dataset and unpaired data separately. Therefore, the final permutation matrix $\left\{\mathbf{P}_{j}\right\}_{j=1}^{v}$ can be fomulated as 
\begin{equation}
    \mathbf{P}_{j}=\left[\begin{array}{cc}
\mathbf{I} & 0 \\
0 & \tilde{\mathbf{P}}_{j} 
\end{array}\right] \text{.}
\end{equation}

Besides, considering the quality of views differs, equally treating the more informative and less informative views will undoubtedly result in the sub-optimal solution. We establish the weight coefficients $\boldsymbol{\alpha}$ by the reconstruction loss of each view, which can be calculated by 
\begin{equation}
\label{opt a}
    \alpha_j = 1 / \left(\mathcal{L}_{con}^{j} \sum_{i = 1}^v{\left(1/{\mathcal{L}_{con}^{i}}\right)}\right),
\end{equation}
where $\mathcal{L}_{con}^{j}$ is reconstruction loss of the $j$-th view  $\boldsymbol{\mathcal{L}}_{con}$.

Finally, we can calculate the fused aligned graph as
\begin{equation}
\label{fuse}
\mathbf{S}_{\text {Aligned}}=\sum_{i=1}^{v} \mathbf{\alpha}_i \mathbf{P}_{i}^\top \mathbf{S}_{i} \mathbf{P}_{i} \text{.}
\end{equation}

The final clustering result can be obtained by simply performing standard spectral clustering on $\mathbf{S}_{\text {Aligned }}$.

\subsection{Implementation}
In this section, we will elaborate on the implementation of how to conduct the proposed UPMGC-SM for establishing the correct cross-view correspondence and getting the fused graph simultaneously in three steps.

\textbf{Step 1. Flexible Graph Generation.} The individual graph from each view can be learned simultaneously by each existing graph-based method. In our implementation, we select three classical graph-based multi-view clustering to validate the effectiveness of our model. 

\textbf{Step 2. Generalized Graph Matching.} After the graph generation, our generalized graph matching is proposed to establish the correct cross-view correspondence.

\textbf{Step 3. Weighted Graph Fusion.} After obtaining the cross-view correspondence, our fused graph can be easily established by solving Eq.\eqref{fuse}

For convenience, the whole process is summarized in Algorithm \ref{algo}.

\begin{algorithm}[t] 

		\caption{Unpaired Multi-View Graph Clustering with Cross-View Structure Matching (UPMGC-SM)}
		\begin{algorithmic}[1]
			\REQUIRE Multi-view dataset  $\left\{ \mathbf{X}_i\right\} _{i=1}^v$, cluster number $k$ 
			\\ \STATE compute $\boldsymbol{\mathcal{L}}_{con}$,$\left\{ \mathbf{S}_i\right\}_{i=1}^v$ by the selected MGC algorithm \\
			\STATE compute $\boldsymbol{\alpha}$ by solving Eq. (\ref{opt a}).
			\FOR{$i$ = 1 : $v$}
			\WHILE{not convergence}
			\STATE compute $\Tilde{\mathbf{P}}_i$ by solving Eq. (\ref{opt P}).
			\ENDWHILE
		   \ENDFOR
			\STATE compute $\mathbf{S}_{\text{Aligned}}$, by solving Eq. (\ref{fuse}).
			\ENSURE Aligned graph $\mathbf{S}_{\text{Aligned}}$
		\end{algorithmic}
  \label{algo}
	\end{algorithm}

\subsection{Theoretical Analysis}
\begin{lemma}\label{lemma 1}
The projection $\boldsymbol{\Gamma}(\cdot)$ is a non-expansive projection on $\boldsymbol{\Delta}$. 
\end{lemma}

\begin{proof}
It is easy to see that $\boldsymbol{\Delta}$ is a convex set, then the $\boldsymbol{\Gamma}(\cdot)$ is a non-expansive projection, which means $ \Vert{\boldsymbol{\Gamma}(\mathbf{S}_1)-\boldsymbol{\Gamma}(\mathbf{S}_2)}\Vert_{F} \leq \Vert\mathbf{S}_1-\mathbf{S}_2\Vert_{F}  $,
which is proven in \cite{sinkhorn1964relationship}. 
\end{proof}

\begin{theorem} \label{theo 1}
If there exists a constant $\varepsilon $ such that $\Vert{\mathbf{S}_{i^*} \otimes \mathbf{S}_{2}}\Vert_{2} \leq \boldsymbol{\varepsilon}$, the algorithm to solve the alignment matrix $\mathbf{P}_i$ converges at rate $1/{2} +\boldsymbol{\varepsilon}$, where $\otimes$ denotes Kronecker product.
\end{theorem}

\begin{proof}
Denote $\mathbf{G}^{(t)} = \mathbf{S}_{2} \mathbf{P}^{(t)} \mathbf{S}_{i^*}^{\top}$. By Lemma \ref{lemma 1}, we have that the projection $\boldsymbol{\Gamma}(\cdot)$ is a non-expansive projection on $\boldsymbol{\Delta}$. Thus, we can easily deduce that 
$\Vert{\boldsymbol{\Gamma}(\mathbf{G}^{(t)})-\boldsymbol{\Gamma}(\mathbf{G}^{(t-1)})}\Vert_{\mathbf{F}} \leq \Vert{\mathbf{G}^{(t)}-\mathbf{G}^{(t-1)}}\Vert_{\mathbf{F}}.$
Then, we have that 
\begin{equation}
\begin{aligned}
&\Vert\mathbf{P}^{(t+1)}-\mathbf{P}^{(t)}\Vert_{\mathbf{F}} \leq \\ &\Vert\mathbf{P}^{(t)}-\mathbf{P}^{(t-1)}\Vert_{\mathbf{F}} / 2+ \Vert\mathbf{G}^{(t)}-\mathbf{G}^{(t-1)}\Vert_{\mathbf{F}} / 2 \nonumber,
\end{aligned}
\end{equation} 
and 
\begin{equation}
\begin{aligned}\Vert\mathbf{G}^{(t)}-\mathbf{G}^{(t-1)}\Vert_{\mathbf{F}} & = 
\Vert\mathbf{S}_{2} \mathbf{P}^{(t)} \mathbf{S}_{i^*}^{\top}-\mathbf{S}_{2} \mathbf{P}^{(t-1)} \mathbf{S}_{i^*}^{\top}\Vert_{\mathbf{F}} \\
&=\Vert\left(\mathbf{S}_{i^*} \otimes \mathbf{S}_{2}\right) \operatorname{vec}\left(\mathbf{P}^{(t)}-\mathbf{P}^{(t-1)}\right)\Vert_{2} \\
&\leq \Vert\left(\mathbf{S}_{2} \otimes \mathbf{S}_{i^*}\right)\Vert_{\mathbf{F}}\Vert\mathbf{P}^{(t)}-\mathbf{P}^{(t-1)}\Vert_{\mathbf{F}} \nonumber \\
& \leq \boldsymbol{\varepsilon} \Vert\mathbf{P}^{(t)}-\mathbf{P}^{(t-1)}\Vert_{\mathbf{F}} \nonumber,
\end{aligned}
\end{equation} 
where the $\operatorname{vec}\left(\cdot \right)$ is the vectorization of a matrix.

Therefore we can obtain that $$\Vert\mathbf{P}^{(t+1)}-\mathbf{P}^{(t)}\Vert_{\mathbf{F}}/{\Vert\mathbf{P}^{(t)}-\mathbf{P}^{(t-1)}\Vert_{\mathbf{F}}} \leq 1/2+\boldsymbol{\varepsilon}.$$
The desirable result follows.
\end{proof}

\subsection{Computational Complexity and Convergence}
\subsubsection{Computational complexity}
 As shown in Algorithm 1, during each iteration, it has three steps: computing $\left\{ \mathbf{S}_i\right\} _{i=1}^v$, computing $\boldsymbol{\alpha}$ and computing $\mathbf{P}$. Without loss of generality, we denote the number of all data as $n$,  the number of unpaired data as $\tilde{n}$, and the number of views as $v$. Obtaining $\left\{ \mathbf{S}_i\right\} _{i=1}^v$ depends on the MSC algorithm selected. We assume the time cost is $\mathcal{O}(f\left(\tilde{n}\right))$. We need to execute $v$ times to get the individual result from each view, so the total cost is $\mathcal{O}(v f\left(n\right))$. Similarly, the time cost of computing $\boldsymbol{\alpha}$ and  $\mathbf{P}$ is $\mathcal{O}(v \tilde{n}^2)$ and $\mathcal{O}(v \tilde{n}^3)$. In total, these three steps cost $\mathcal{O}(v \left(f\left(n\right) + \tilde{n}^2 + \tilde{n}^3 \right))$ time, where $v$ is the number of views. At last, the time consumption of the standard clustering is $\mathcal{O}(f\left(n\right) + \tilde{n}^3)$. Taking the three methods selected in this paper, i.e., LSR, GMC, and CoMSC, their algorithm complexity is $\mathcal{O}(n^3)$, $\mathcal{O}(((m k+m n+c+c n) n) t+m n k d)$, and $\mathcal{O}(tmn^3)$, respectively. Because of that, the time complexity of the above three methods combined with our proposed framework is $\mathcal{O}(v(n^3 + \tilde{n}^3))$,  $\mathcal{O}(v(((m k+m n+c+c n) n) t+m n k d)+\tilde{n}^3)$,  and $\mathcal{O}(v(tmn^3)+\tilde{n}^3)$, respectively.
    
    \subsubsection{Convergence}	
   By following Theorem \ref{theo 1}, we can get that our algorithm is convergent.

\section{Experiment}
In this section, we conduct comprehensive experiments to evaluate the effectiveness of our proposed UPMGC-SM. Specifically, the clustering performance on fully and partially unpaired data, the evolution of the objective value, and view coefficients are carefully discussed. Our code is publicly available at \url{https://github.com/wy1019/UPMGC-SM}.

\begin{table}[t]
\caption{Benchmark datasets}
\renewcommand\arraystretch{1.38}
\tabcolsep=0.1cm
	\centering
    \scalebox{1.05}{
\begin{tabular}{ccccc}
\toprule
\multirow{2}{*}{Dataset} & \multicolumn{4}{c}{Number of}                       \\ \cline{2-5} 
                         & Samples & Views & Clusters & Features               \\ \hline
3Sources                 & 169     & 3     & 6        & 3560,3631,3068         \\
BBCSport3                & 282     & 3     & 5        & 2582,2544,2465         \\
ORL                      & 400     & 3     & 40       & 4096,3304,6750         \\
Prokaryotic              & 551     & 3     & 4        & 438,3,393              \\
WebKB                    & 1051    & 2     & 2        & 2949,334               \\
Caltech101-7             & 1474    & 6     & 7        & 48,40,254,1984,512,928 \\
HandWritten              & 2000    & 6     & 10       & 216,76,64,6,240,47     \\
Wiki                     & 2866    & 2     & 10       & 128,10                 \\
Hdigit                   & 10000   & 2     & 10       & 784,256                \\ \bottomrule
\end{tabular}}
\label{benchmark_data}
\end{table}
	\vspace{7pt}

\subsection{Datasets}
Nine wide-used datasets are adopted to evaluate the effectiveness of the proposed algorithm. The details of the datasets are as follows, including 3Sources, BBCSport3, ORL, Prokaryotic, WebKB, Caltech101-7, HandWritten,  Wiki, and Hdigit. The detailed information is listed in Table \ref{benchmark_data}.

\textbf{3Sources}:\footnote{\url{http://mlg.ucd.ie/datasets/3sources.html}} It is a dataset of 169 news articles from three well-known online news sources, where each source is considered as one view.

\textbf{BBCSport3}:\footnote{\url{http://mlg.ucd.ie/datasets/segment.html}} It contains 544 new sports articles collected from five topical areas, which correspond to five classes (athletics, cricket, football, rugby, and tennis). The document is described by three views, and their dimensions are 2582, 2544 and 2465 respectively.

\textbf{ORL}:\footnote{\url{https://cam-orl.co.uk/facedatabase.html}} It contains 400 facial images of 40 people at different times, lighting, and facial details. The three feature sets are 4096-dimensional intensity, 3304-dimensional LBP, and 6750-dimensional Gabor.

\textbf{Prokaryotic}:\footnote{\url{https://github.com/mbrbic/Multi-view-LRSSC/tree/master/datasets}} contains multiple prokaryotic species describing heterogeneous multi-view data, including text and different genomic representations.

\textbf{WebKB}:\footnote{\url{http://lig-membres.imag.fr/grimal/data.html}} It is described by two aspects of the web page, \textit{i.e.}, content and links. They were collected from four universities, and the state of Wisconsin was chosen.

\textbf{Caltech101-7}:\footnote{\url{https://data.caltech.edu/records/mzrjq-6wc02}} It is a subset of Caltech101, which collects 1474 images of objects belonging to 101 categories.
It is a collection of a large number of images of objects belonging to 101 categories. Among them, nine popular categories have been selected, including Human Face, Motorcycle, Dollar Bill, Garfield, Snoopy, Parking Sign, and Windsor Chair.

\textbf{Handwritten}:\footnote{\url{https://archive.ics.uci.edu/ml/datasets/Multiple+Features}} It includes 2000 handwritten digital images from 0 to 9. Each sample is represented by six different feature sets, namely 216-dimensional FAC, 76-dimensional FOU,  64-dimensional KAR, 6 MORs, 240-dimensional Pix, and 47-dimensional ZER.

\textbf{Wiki}:\footnote{\url{http://www.svcl.ucsd.edu/projects/crossmodal/}} It consists of 2866 sections selected from Wikipedia's collection of featured articles, where text and SIFT histograms are used for text and images, respectively.

\textbf{Hdigit}: \footnote{\url{https://cs.nyu.edu/∼roweis/data.html}} This handwritten digits (0-9) data set is from two sources,\textit{i.e.}, MNIST Handwritten Digits and USPS Handwritten Digits. The data set consists of 10000 samples.

\subsection{Evaluation Metrics}
To evaluate the clustering performance, three widely used metrics, including accuracy (ACC), normalized mutual information (NMI), and purity \cite{CCGC,GCC-LDA}, are adopted to verify the clustering performance. Some figures are only presented on several datasets, and others are given in the appendix due to space limits. 
 For $k$-means clustering evaluation, we repeat this process for $50$ times and record their average values as final clustering results. All the experiments are conducted on a desktop computer with Intel(R) Core(TM)-i7-7820X CPU and 64G RAM.
\begin{table*}[t]
\caption{Empirical evaluation and comparison of UPMGC-SM with nine baseline methods on nine unpaired benchmark datasets in terms of ACC, NMI, and Purity}
\label{results}
\selectfont 
\centering
\renewcommand\arraystretch{1.17}
\tabcolsep=0.1cm
\scalebox{1}{
\begin{tabular}{cccccccccc}
\toprule
Algorithm  & 3Sources                          & BBCSport3                         & ORL                               & Prokaryotic                       & WeBKB                             & Caltech101-7                      & HandWritten                       & Wiki                              & Hdigit                            \\ \hline
\multicolumn{10}{c}{ACC (\%)}                                                                                                                                                                                                                                                                                                                  \\ \hline
LSR \cite{lu2012robust}    & 31.86±2.54                        & 35.95±1.98                        & 41.25±2.03                        & 32.60±3.71                        & 74.21±1.20                        & 17.26+2.23                        & 40.09±2.31                        & 14.88±1.03                        & 11.43±1.06                        \\
GMC \cite{wang2019gmc}    & 45.56±2.82                        & 35.11±1.44                        & 11.82±1.31                        & 38.66±2.21                        & 77.83±1.18                        & 18.82±1.43                        & 30.22±1.94                        & 36.17±2.47                        & 16.51±2.78                        \\
CoMSC \cite{liu2021multiview}  & 41.44±4.45                        & 43.76±2.50                        & 34.76±1.81                        & 45.94±2.35                        & 69.64±3.17                        & 23.14±1.01                        & 16.91±2.46                        & 19.65±1.40                        & 33.68±2.41                        \\
LMSC  \cite{zhang2017latent}   & 36.90±1.51                        & 34.81±3.57                        & 17.34±2.70                        & 41.08±2.49                        & 34.81±2.57                        & 25.16±2.27                        & 13.49±2.15                        & 17.31±2.10                        & 29.19±3.54                        \\
MVEC \cite{tao2017ensemble}   & 39.83±2.23                     &  37.08±1.51                      &  17.29±0.65                     &  41.35±0.09                       &   75.45±0.00                      & 22.09±1.83                     &   22.58±0.14                        & 24.65±0.05                      & 30.33±0.22                    \\
CSMSC \cite{luo2018consistent}  & 34.02±2.52                        & 30.22±2.94                        & 16.91±2.46                        & 33.66±1.71                        & 75.36±1.36                        & 24.78±2.84                        & 13.06±2.08                        & 16.41±2.14                        & 11.27±3.01                        \\
CGL \cite{li2021consensus}    & 33.11±2.45                        & 36.17±2.30                        & 19.65±1.40                        & 43.19±3.01                        & 65.56±1.33                        & 12.29±2.34                        & 16.61±2.12                        & 13.38±2.22                        & 18.81±2.78                        \\
OPMC \cite{liu2021one}   & 37.81±1.31                        & 35.53±0.45                        & 30.25±0.58                        & 53.95±0.86                        & 55.57±0.98                        & 25.24±1.52                        & 23.10±0.45                        & 27.93±0.14                        & 43.80±1.08                        \\ \hline
PVC$^{+}$ \cite{huang2020partially}    & 33.97±2.71                        & 36.57±1.75                        & 37.57±2.75                        & {\color[HTML]{4472C4} 56.27±2.28} & 78.41±2.35                        & {\color[HTML]{4472C4} 47.75±2.20} & 28.05±2.93                        & 15.95±2.42                        & 36.04±2.35                        \\
MvCLN$^{+}$ \cite{yang2021partially} & 30.03±3.99                        & 32.34±2.41                        & 30.20±2.45                        & 50.20±9.33                        & 75.55±4.01                        & 36.16±3.14                        & 69.00±3.90                        & 50.55±3.95                        & O/M                               \\ \hline
MVC-UM$^{*}$ \cite{yu2021novel} & 34.91±2.11                        & 34.75±0.17                        & 58.95±2.63                        & 53.05±4.45                        & 62.62±3.45                        & 38.74±3.77                        & {\color[HTML]{4472C4} 69.02±3.52} & {\color[HTML]{4472C4} 54.42±1.65} & O/M                               \\ \hline
LSR+Ours   & 48.21±2.83                        & 36.97±0.33                        & {\color[HTML]{4472C4} 71.60±0.57} & 53.97±0.47                        & 76.02±0.00                        & 37.97±0.41                        & {\color[HTML]{FF0000} 70.70±1.18} & 53.39±0.30                        & {\color[HTML]{4472C4} 53.29±0.07} \\
GMC+Ours   & {\color[HTML]{4472C4} 55.53±3.62} & {\color[HTML]{4472C4} 47.18±1.70} & 69.71±1.20                        & 46.07±0.11                        & {\color[HTML]{4472C4} 80.59±0.00} & 33.54±0.66                        & 44.55±0.47                        & {\color[HTML]{FF0000} 55.14±0.23} & 46.60±0.23                        \\
CoMSC+Ours & {\color[HTML]{FF0000} 63.18±4.63} & {\color[HTML]{FF0000} 56.59±1.98} & {\color[HTML]{FF0000} 76.24±3.18} & {\color[HTML]{FF0000} 57.65±4.19} & {\color[HTML]{FF0000} 83.36±4.60} & {\color[HTML]{FF0000} 54.21±0.73} & {\color[HTML]{4472C4} 70.08±1.35} & {\color[HTML]{4472C4} 54.67±1.72} & {\color[HTML]{FF0000} 63.35±1.35} \\ \hline
\multicolumn{10}{c}{NMI (\%)}                                                                                                                                                                                                                                                                                                                  \\ \hline
LSR \cite{lu2012robust}    & 17.32±1.86                        & 10.17±1.06                        & 62.07±1.37                        & 12.00±2.20                        & 3.01±2.02                         & 11.01+2.27                        & 28.76±1.84                        & 12.27±1.02                        & 5.19±1.03                         \\
GMC \cite{wang2019gmc}    & 19.78±2.37                        & 7.89±1.67                         & 27.90±1.24                        & 9.28±2.76                         & 0.26±1.17                         & 10.80±2.04                        & 1.52±1.67                         & 7.89±2.88                         & 3.71±2.61                         \\
CoMSC \cite{liu2021multiview}  & 22.92±2.94                        & 17.01±2.83                        & 57.23±1.08                        & 1.36±2.10                         & {\color[HTML]{4472C4} 26.69±3.28} & 20.53±2.01                        & 0.51±2.74                         & 27.90±1.24                        & 22.01±2.28                        \\
LMSC  \cite{zhang2017latent}   & 21.40±1.90                        & 12.43±2.75                        & 41.60±2.48                        & 8.13±2.92                         & 12.43±2.75                        & 21.77±2.19                        & 1.28±1.86                         & 9.28±1.86                         & 20.91±2.97                        \\
MVEC \cite{tao2017ensemble}   & 13.64±0.56                        & 2.95±0.21                         & 41.23±0.51                        & 13.54±0.03                        & 0.12±0.00                         & 4.96±1.55                         & 10.67±0.04                        & 12.09±0.03                        & 21.41±0.09                        \\
CSMSC \cite{luo2018consistent}  & 11.91±1.53                        & 5.20±1.99                         & 41.02±1.39                        & 6.56±1.35                         & 0.12±2.06                         & 10.63±3.04                        & 0.98±3.02                         & 0.26±1.07                         & 3.01±2.77                         \\
CGL \cite{li2021consensus}    & 20.39±2.45                        & 14.01±2.03                        & 35.25±1.32                        & 12.51±2.01                        & 1.07±3.01                         & 13.13±2.59                        & 3.04±1.84                         & 1.52±2.18                         & 10.99±3.12                        \\
OPMC \cite{liu2021one}   & 16.19±0.66                        & 11.66±0.49                        & 42.48±1.06                        & 31.01±1.40                        & 5.73±1.40                         & 20.76±0.69                        & 10.76±0.72                        & 13.99±0.47                        & 41.70±0.31                  \\ \hline
PVC$^{+}$ \cite{huang2020partially}    & 19.96±2.83                        & 11.18±0.28                        & 64.07±1.59                        & {\color[HTML]{4472C4} 31.58±2.55} & 10.06±2.92                        & 23.35±2.29                        & 31.51±2.76                        & 3.53±2.38                         & 19.60±2.57                        \\
MvCLN$^{+}$ \cite{yang2021partially}  & 10.20±3.47                        & 4.86±1.58                         & 57.89±3.38                        & 20.50±4.44                        & 1.89±3.27                         & 16.44±3.05                        & {\color[HTML]{FF0000} 68.77±3.85} & 39.25±4.58                        & O/M                               \\ \hline
MVC-UM$^{*}$ \cite{yu2021novel} & 6.22±2.08                         & 2.35±0.41                         & 47.67±2.07                        & 27.02±2.46                        & 9.70±1.86                         & {\color[HTML]{4472C4} 27.87±3.03} & 17.32±1.86                        & 19.78±0.37                        & O/M                               \\ \hline
LSR+Ours   & 32.65±2.62                        & 3.67±1.53                         & {\color[HTML]{4472C4} 86.72±0.28} & 30.58±0.32                        & 5.67±0.79                         & 26.23±0.39                        & {\color[HTML]{4472C4} 66.70±0.60} & 51.93±0.18                        & {\color[HTML]{4472C4} 45.35±0.04} \\
GMC+Ours   & {\color[HTML]{4472C4} 34.26±2.29} & {\color[HTML]{4472C4} 19.71±1.63} & 84.55±0.73                        & 25.48±0.12                        & 14.84±0.00                        & 21.69±2.53                        & 49.30±0.17                        & {\color[HTML]{FF0000} 54.07±0.05} & 44.33±0.11                        \\
CoMSC+Ours & {\color[HTML]{FF0000} 61.36±2.94} & {\color[HTML]{FF0000} 46.77±1.80} & {\color[HTML]{FF0000} 92.20±0.97} & {\color[HTML]{FF0000} 32.88±0.91} & {\color[HTML]{FF0000} 31.71±2.93} & {\color[HTML]{FF0000} 52.19±1.00} & 62.44±0.29                        & {\color[HTML]{4472C4} 53.68±0.91} & {\color[HTML]{FF0000} 57.04±0.69} \\ \hline
\multicolumn{10}{c}{Purity (\%)}                                                                                                                                                                                                                                                                                                               \\ \hline
LSR \cite{lu2012robust}    & 49.73±1.86                        & 44.60±1.97                        & 44.46±2.85                        & 56.81±2.10                        & 72.34±2.24                        & 54.13+2.29                        & 40.68±1.38                        & 17.59±1.03                        & 11.57±1.10                        \\
GMC \cite{wang2019gmc}    & 46.75±1.31                        & 37.23±2.77                        & 14.39±1.27                        & 56.81±2.68                        & 78.12±1.00                        & 54.14±1.57                        & 10.65±1.48                        & 16.36±2.22                        & 18.81±2.34                        \\
CoMSC \cite{liu2021multiview}  & 54.60±2.27                        & 48.35±0.60                        & 37.36±1.82                        & 56.81±2.01                        & 69.64±2.07                        & 24.20±2.01                        & 20.43±2.85                        & 20.42±1.19                        & 34.03±2.67                        \\
LMSC  \cite{zhang2017latent}   & 54.96±1.13                        & 48.96±2.63                        & 18.11±2.77                        & 56.81±2.00                        & 48.96±1.63                        & 54.87±2.26                        & 14.02±2.21                        & 19.50±2.16                        & 31.59±3.05                        \\
MVEC \cite{tao2017ensemble}   & 44.90±0.78                        & 38.29±0.56                        & 18.06±0.75                        & 57.60±0.00                        & 78.12±0.00                        & 59.99±2.29                        & 22.83±0.15                        & 27.09±0.03                        & 30.89±0.09                        \\
CSMSC \cite{luo2018consistent}  & 40.26±2.12                        & 36.99±2.10                        & 17.85±1.51                        & 56.81±1.22                        & 78.12±1.82                        & 54.14±2.77                        & 13.35±3.10                        & 18.92±0.91                        & 15.11±2.81                        \\
CGL \cite{li2021consensus}    & 47.28±0.50                        & 49.65±2.77                        & 19.69±1.39                        & 56.81±2.67                        & 78.12±3.29                        & 55.34±2.56                        & 17.58±2.13                        & 16.25±2.07                        & 19.93±2.82                        \\
OPMC \cite{liu2021one}  & 52.60±1.04                        & 46.52±0.68                        &  31.50±0.67                        &  63.67±1.80                        &  78.12±1.51                        &  54.14±1.59                        &  23.30±0.39                       &  29.82±0.52                       & 48.26±1.25                    \\ \hline
PVC$^{+}$ \cite{huang2020partially}    & 39.07±2.08                        & 42.55±0.18                        & 42.16±2.05                        & 70.59±2.53                        & 79.73±2.78                        & 73.53±2.93                        & 37.89±2.13                        & 20.65±2.47                        & 36.57±2.01                        \\
MvCLN$^{+}$ \cite{yang2021partially}  & 40.83±4.14                        & 39.36±2.84                        & 32.90±2.90                        & 61.96±5.15                        & 78.12±4.82                        & 61.47±3.23                        & 69.84±3.09                        & 53.83±5.12                        & O/M                               \\ \hline
MVC-UM$^{*}$ \cite{yu2021novel} & 35.03±2.01                        & 40.15±0.68                        & 47.67±2.07                        & 63.18±4.63                        & 78.20±0.59                        & 68.55±3.29                        & 17.32±1.86                        & 19.78±3.01                        & O/M                               \\ \hline
LSR+Ours   & 59.80±3.73                        & 42.26±0.39                        & {\color[HTML]{4472C4} 75.58±0.47} & 62.96±0.12                        & 78.12±0.00                        & 66.33±0.24                        & 72.04±0.84                        & 60.50±0.25                        & {\color[HTML]{4472C4} 55.29±0.02} \\
GMC+Ours   & {\color[HTML]{4472C4} 68.03±1.14} & {\color[HTML]{4472C4} 55.57±0.99} & 73.32±1.09                        & 64.38±0.13                        & {\color[HTML]{4472C4} 80.59±0.00} & 66.74±1.66                        & 47.68±0.12                        & {\color[HTML]{FF0000} 61.97±0.05} & 49.11±0.18                        \\
CoMSC+Ours & {\color[HTML]{FF0000} 75.89±1.97} & {\color[HTML]{FF0000} 63.16±1.28} & {\color[HTML]{FF0000} 81.07±2.23} & {\color[HTML]{FF0000} 69.77±1.56} & {\color[HTML]{FF0000} 83.90±0.83} & {\color[HTML]{FF0000} 82.02±1.01} & {\color[HTML]{FF0000} 69.98±0.18} & {\color[HTML]{4472C4} 61.74±1.36} & {\color[HTML]{FF0000} 64.03±2.60} \\ \bottomrule
\end{tabular}}
\end{table*}

\subsection{Compared Methods}
Along with our proposed UPMGC-SM, We run eleven state-of-the-art multi-view clustering methods for comparison. Specifically, eight compared methods are designed for fully paired data, two for partially unpaired data, and one for fully unpaired data. The details of them are provided as follows.
\subsubsection{Multi-View Clustering Algorithms on Fully Paired Data}
\begin{itemize}
	\item {\bf Least squares regression (LSR)} \cite{lu2012robust}:  This method performs subspace clustering on each view and simply contacts all views into a single one.
	\item {\bf Graph-based multi-view clustering (GMC)} \cite{wang2019gmc}: This method learns a common graph by combining adaptive neighbor graph learning and graph fusion in a unified framework.
	\item {\bf Multi-view subspace clusering via co-training robust data representation (CoMSC)} \cite{liu2021multiview}: This method performs the multi-view subspace with a discriminative data representation which is generated by the eigen decomposition.
	\item {\bf Latent multi-view subspace clustering (LMSC)} \cite{zhang2017latent}: This method integrates latent representation learning and multi-view subspace clustering in a unified framework to learn a common subspace.
\item {\bf Multi-View Ensemble Clustering (MVEC)}\cite{tao2017ensemble} This method leverages the complementary information of multi-view data by generating Basic Partitions (BPs) for each view individually and seeking a consensus partition among all the BPs.
	\item {\bf Consistent and specific multi-view subspace clustering (CSMSC)}  \cite{luo2018consistent}: This method simultaneously learns the view-consistent representation and the view-specific representations to generate the common subspace.
	\item {\bf Consensus graph learning for multi-view clustering (CGL)} \cite{li2021consensus}: This method generates the similarity graph by integrating the spectral embedding and low-rank tensor learning into a unified framework.
\item {\bf One-pass multi-view clustering for large-scale data (OPMC)} \cite{liu2021one}: This method proposes an one-pass multi-view clustering method by removing the non-negativity constraint and jointly optimize the aforementioned two steps.
\end{itemize}
\begin{figure*}[t]
\begin{center}{
    \centering
     \begin{minipage}[t]{0.95\linewidth}
        \centering
        \subfloat{
            \includegraphics[width=0.99\linewidth]{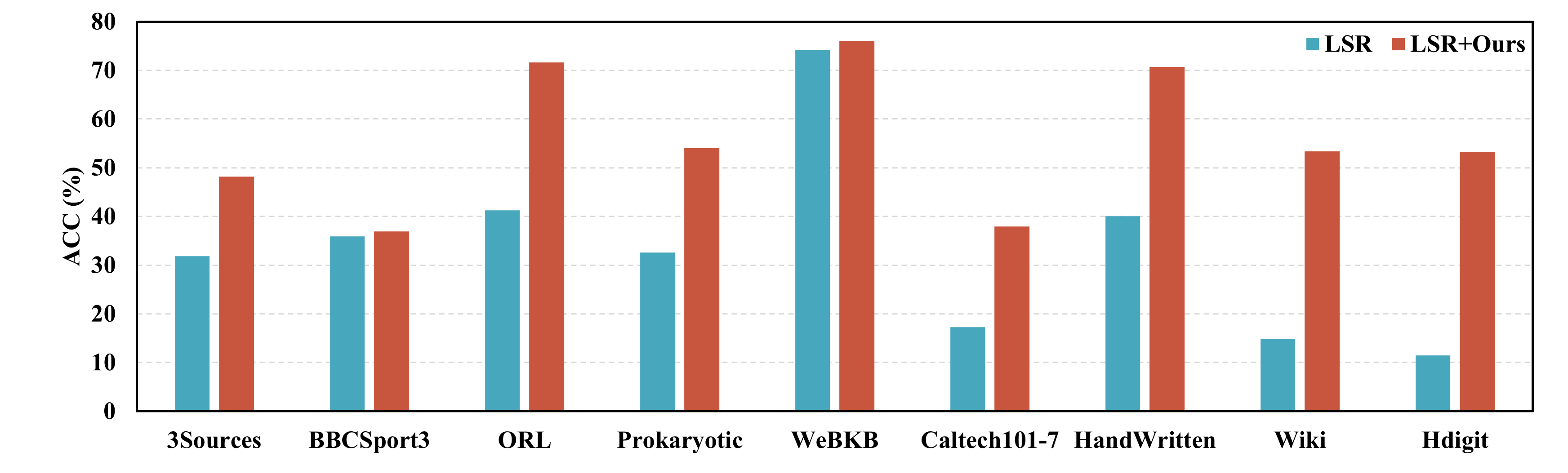}
        }

        \subfloat{
            \includegraphics[width=0.99\linewidth]{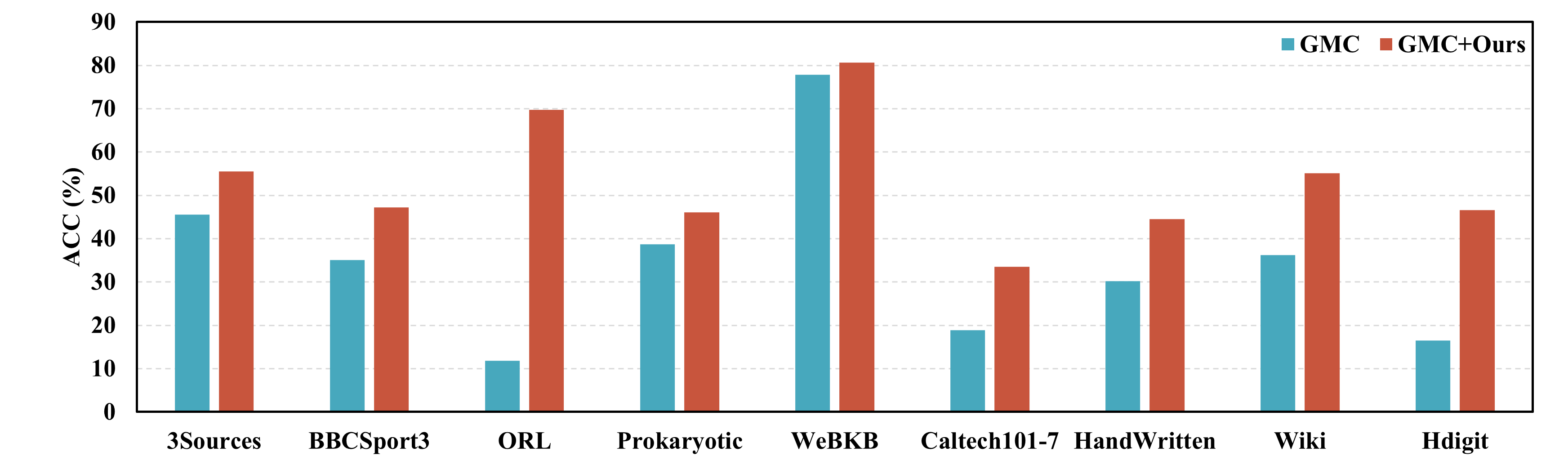}}

        \subfloat{
            \includegraphics[width=0.99\linewidth]{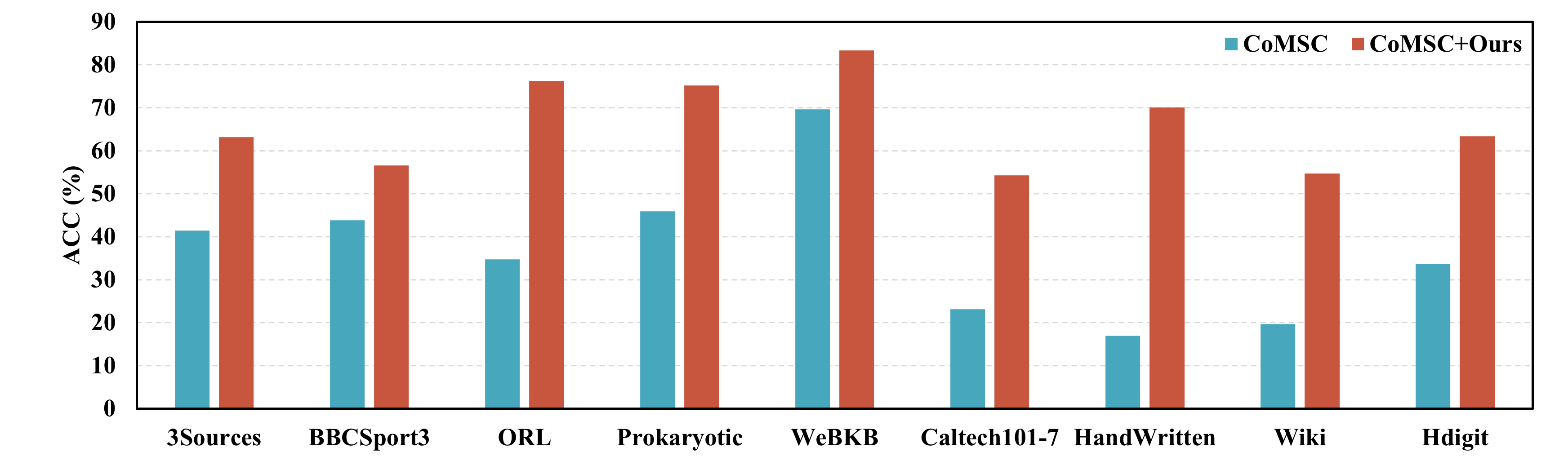}
        }
    \end{minipage}}
    \end{center}
    \caption{Performance comparison w./w.o. the UPMGC-SM framework on nine datasets.}
    \label{fig:result_compar}
    \vspace{-0.28 cm}
\end{figure*}
\vspace{7pt}
\subsubsection{Multi-View Clustering Algorithm on Partially Unpaired  data}

\begin{itemize}
	\item {\bf Partially view-aligned clustering (PVC)} \cite{huang2020partially}: This method uses a differentiable surrogate of the Hungarian algorithm to learn the correspondence of the unaligned data in a latent space learned by a neural network 
 
    \item{\bf Multi-view Contrastive
Learning with Noise-robust Loss (MvCLN)} \cite{yang2021partially}:
 This method proposes a novel noise-robust contrastive loss to establish the view correspondence using contrastive
learning
\end{itemize}

\subsubsection{Multi-View Clustering Algorithm on Fully Unpaired Data}
\begin{itemize}
	\item {\bf Multi-view clustering for unknown mappings (MVC-UM)} \cite{yu2021novel}: This method builds reconstruction error terms, local structural constraint terms, and cross-view mapping loss to establish the correspondence of cross-view data based on the non-negative matrix factorization.
\end{itemize}

For the above algorithms, we directly use the code of the comparison method from the author's website to perform a grid search in the set of parameters recommended in his paper and report the best results.
\begin{figure*}[t]

\centering
\begin{minipage}[t]{0.6\linewidth}
\centering
\subfloat{
    \includegraphics[width=0.99\linewidth]{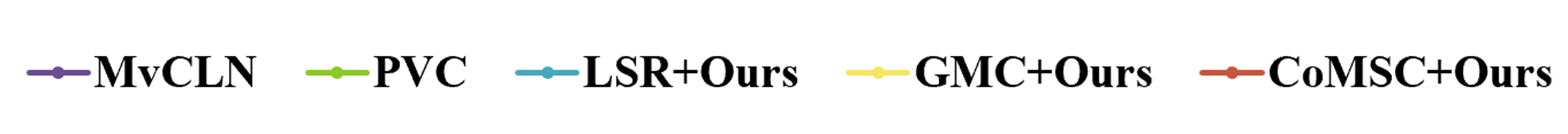}
}
\end{minipage}
\qquad
\begin{minipage}[t]{0.99\linewidth}
\centering
\subfloat{
    \includegraphics[width=1\linewidth]{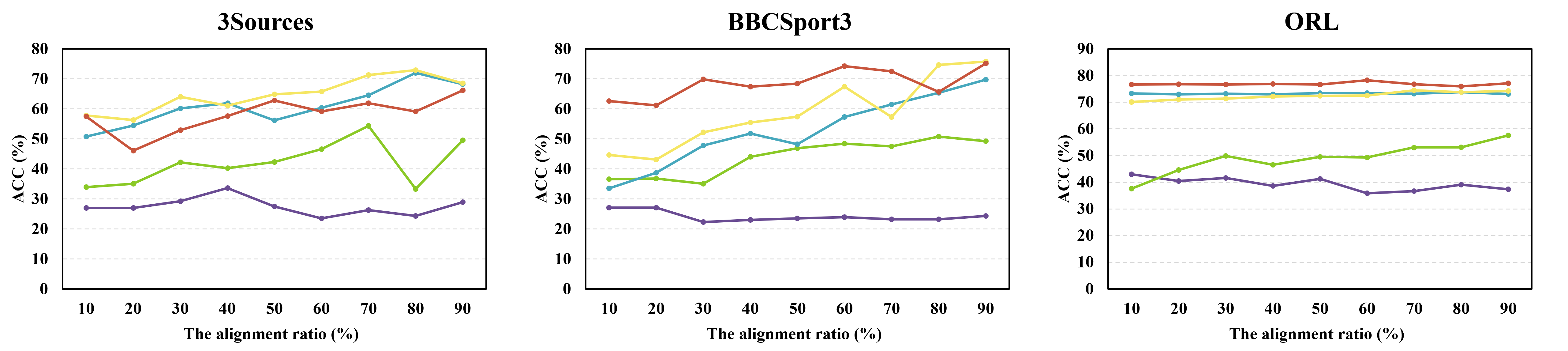}
}
\end{minipage}
\qquad
\begin{minipage}[t]{0.99\linewidth}
    \centering
    \subfloat{
        \includegraphics[width=1\linewidth]{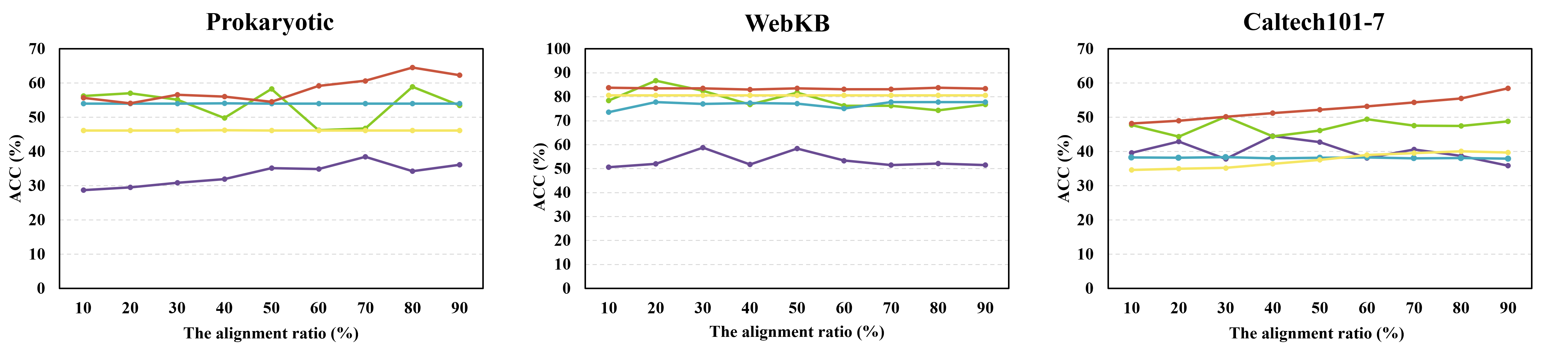}
    }
\end{minipage}
\qquad
\begin{minipage}[t]{0.99\linewidth}
    \centering
    \subfloat{
        \includegraphics[width=1\linewidth]{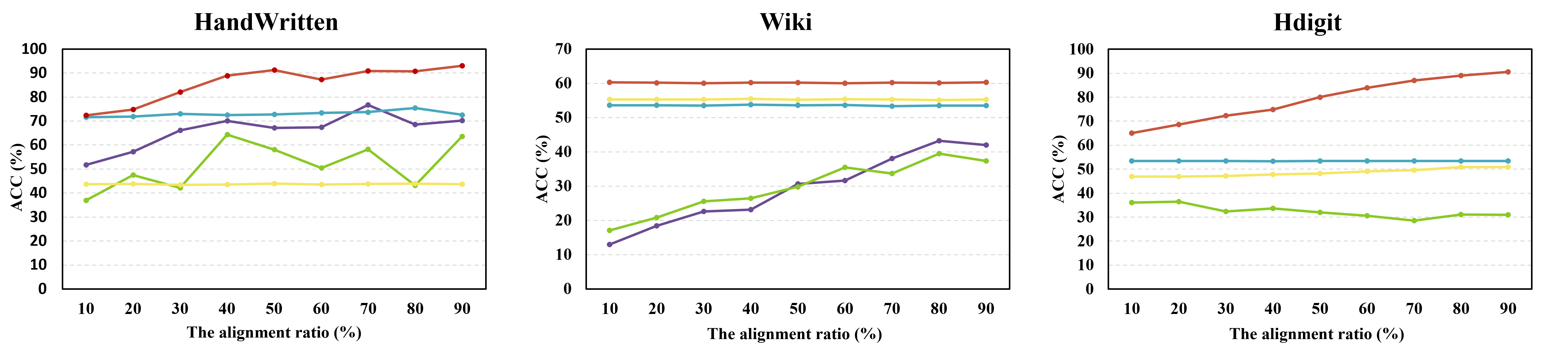}
    }
\end{minipage}

\caption{Clustering performance
of UPMGC-SM on nine datasets with varying alignment ratio.}
\label{par_result}
\end{figure*}

\subsection{Fully Unpaired Data Construction}
To evaluate the performance of UPMGC-SM on fully unpaired data, we construct the fully unpaired data by setting the first view as the baseline view and randomly disrupting the samples in other views. Considering PVC and MvCLN can only handle the partially unpaired data,  we remain 10\% of the paired data to adapt to the unpaired setting.

For each experiment on the fully unpaired data set, we randomly disturb datasets 10 times and report the means and variances of their results to eliminate the effect of randomness on our results.

\subsection{Experimental Results on Fully Unpaired Data}

Table \ref{results} reports the clustering results on nine benchmark datasets. We mark the optimal results in red, while the sub-optimal and close results are in blue, and 'O/M' indicates out-of-memory issue. In the comparison experiments, we labeled the algorithms that can handle partially unpaired data and fully unpaired data with $+$ and $*$, respectively. It is worth noting that for those algorithms that can only handle fully paired data, the different ground truths of each view will have an impact on their clustering results. In this experimental species, we uniformly use the label of its undisturbed first view as the evaluation criterion for its final clustering results.

\begin{figure*}
    \centering
    \includegraphics[height = 85pt, width=0.99\linewidth]{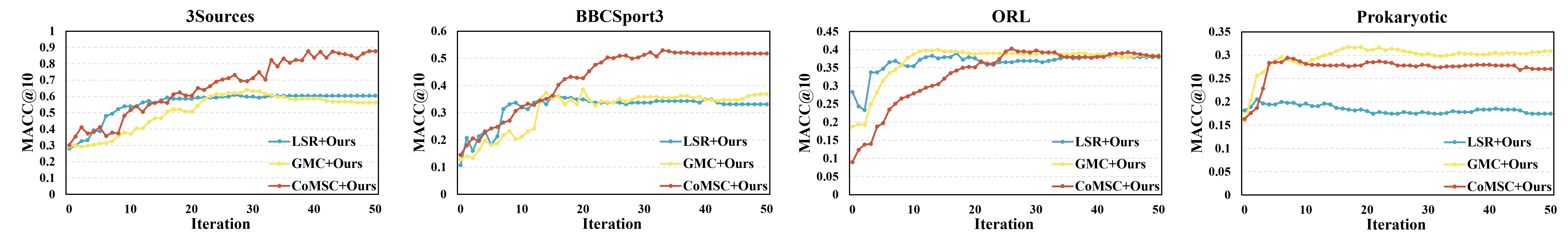}
    \caption{Variation of MACC@10 on four datasets with increasing number of iterations}
    \label{macc}
\end{figure*}
\vspace{7pt}
\begin{figure*}
    \centering
    \includegraphics[height = 85pt ,width=0.99\linewidth]{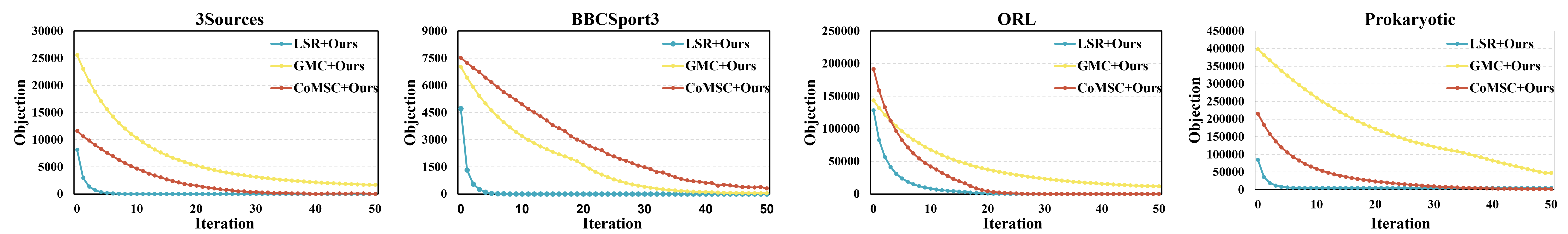}
    \caption{Variation of objection on four datasets with increasing number of iterations}
    \label{objection}
\end{figure*}
\vspace{7pt}
  \begin{table*}[t]
\caption{Comparing the effectiveness of our selection principle (ACC (\%)). ‘Unchosen’ denotes simply selecting the first view of each dataset to align. ‘Chosen’ denotes alignment by our selection principle.}
\centering
\renewcommand\arraystretch{1.25}
\scalebox{1}{
\resizebox{\linewidth}{!}{
\begin{tabular}{ccccccccccc}
\toprule
\multicolumn{2}{c}{Method}         & 3Sources                     & BBCSport                     & ORL                          & Prokaryotic                  & WebKB                        & Caltech101-7                 & HandWritten                  & Wiki                         & Hdigit                       \\ \hline
                        & Unchosen & 37.73                        & 28.40                        & 66.31                        & 31.58                        & 58.04                        & 35.43                        & 12.90                        & 13.48                        & 50.55                        \\
\multirow{-2}{*}{LSR}   & Chosen   & {\color[HTML]{FF0000} 48.21} & {\color[HTML]{FF0000} 36.97} & {\color[HTML]{FF0000} 71.60} & {\color[HTML]{FF0000} 53.97} & {\color[HTML]{FF0000} 76.02} & {\color[HTML]{FF0000} 37.97} & {\color[HTML]{FF0000} 70.70} & {\color[HTML]{FF0000} 26.75} & {\color[HTML]{FF0000} 53.29} \\ \hline
                        & Unchosen & 43.50                        & 38.89                        & 46.24                        & 39.16                        & 68.32                        & {\color[HTML]{FF0000} 36.54} & 40.86                        & 23.89                        & {\color[HTML]{FF0000} 52.77} \\
\multirow{-2}{*}{GMC}   & Chosen   & {\color[HTML]{FF0000} 55.53} & {\color[HTML]{FF0000} 47.18} & {\color[HTML]{FF0000} 69.71} & {\color[HTML]{FF0000} 46.07} & {\color[HTML]{FF0000} 80.59} & 33.54                        & {\color[HTML]{FF0000} 44.55} & {\color[HTML]{FF0000} 50.01} & 46.60                        \\ \hline
                        & Unchosen & {\color[HTML]{FF0000}66.51 }                       & 43.91                        & 68.25                        &  55.62 & 77.35                        & 50.58                        & {\color[HTML]{FF0000} 76.66} & 18.39                        & 61.37                        \\
\multirow{-2}{*}{CoMSC} & Chosen   &  63.18 & {\color[HTML]{FF0000} 56.59} & {\color[HTML]{FF0000} 76.24} & {\color[HTML]{FF0000}57.65}                        & {\color[HTML]{FF0000} 83.36} & {\color[HTML]{FF0000} 54.21} & 70.08                        & {\color[HTML]{FF0000} 54.66} & {\color[HTML]{FF0000} 63.35} \\
\bottomrule
\end{tabular}}}
\label{selection}
\end{table*}
\vspace{7pt}
According to the results, we have the following conclusions

\begin{itemize}
	\item In general, our proposed framework outperforms the other nine comparison methods on most of the data sets, which demonstrates the effectiveness of our approach in handling fully unpaired data.
    \item  As the classical graph clustering algorithms, CSMSC and CGL show poor clustering performance on fully unpaired datasets, which illustrates the incapability of most existing methods to handle fully unpaired scenarios because the incorrect alignment information between views will mislead the final clustering partition.
	\item CoMSC achieves better performance than MVC-UM on 3Sources and BBCSport3, but we must emphasize: the clustering performance of CoMSC varies significantly on different views, which indicates that traditional  MVC methods can also handle fully unpaired data in some specific cases. Besides, we can observe: although CoMSC achieves relatively superior performance on 3Sources and BBCSport3, it still achieves a considerable improvement when combined with our approach.
 \item Compared to PVC and MvCLN, which can handle partially unpaired data, UPMGC-SM achieves better results on the majority of datasets. specifically, with respect to ACC, our algorithm LSR+Ours exceeds MvCLN on nine datasets with \textbf{18.18}\%, \textbf{4.63}\%, \textbf{41.40}\%, \textbf{3.77}\%, \textbf{0.47}\%,  \textbf{1.81 }\%, 
  \textbf{1.70 }\%, \textbf{2.84}\%, \textbf{53.29}\%, which indicates that existing multi-view clustering algorithms dealing with partially paired data cannot handle low-alignment scenarios because the lower alignment information cannot provide valid and learnable alignment relationships.
  \item Compared to MVC-UM, which can also handle fully unpaired data, UPMGC-SM achieves great superiority on most datasets. Specifically,  in terms of ACC, CoMSC+Ours outperforms MVC-UM on nine datasets with \textbf{28.27}\%, \textbf{21.84}\%, \textbf{17.29}\%, \textbf{22.09}\%, \textbf{20.74}\%,  \textbf{15.47}\%, 
  \textbf{1.06}\%, \textbf{0.25}\%, \textbf{17.34}\%, which we conjecture that it may be because of the fact that MVC-UM only selects information from a single view of the data for clustering, but does not effectively fuse information from multiple views, which also demonstrates the superiority of our framework.
	\end{itemize}

In Figure \ref{fig:result_compar}, we show the results on the ACC obtained for the ablation studies of LSR, GMC, and CoMSC w./ w.o. our framework. It can be found that for the three algorithms mentioned above on all nine benchmarks. The clustering performance obtained using the alignment module is better than its original value.

\begin{itemize}

	\item As a weak baseline, LSR \cite{lu2012robust} achieves poor performance in comparison with most MVC algorithms, as shown in Table \ref{results}. However, by combining our framework, its performance improves significantly and consistently outperforms the second-best algorithm MVC-UM \cite{yu2021novel} on five datasets in terms of ACC on the benchmark datasets.
	\item For all three baseline methods, our model can significantly improve their performance, which verifies the effectiveness of the proposed method. Specifically, for all the benchmark datasets, CoMSC \cite{liu2021multiview} with our framework  achieves about {\bf 22.74\%}, {\bf 12.83\%}, {\bf 41.48\%}, {\bf 29.2\%}, {\bf 3.72\%}, {\bf 50.56\%}, {\bf 37.28\%} on ACC compared to the original results.
     \item There are differences in the performance improvement of our framework on different datasets, and we conjecture that the number of clusters and views has an impact on the final cluster results. As is shown in Table \ref{benchmark_data}, WebKB only has two samples and two views, while HandWritten has ten clusters and six views. With fewer clusters and views, WebKB samples have a higher probability of being assigned to the correct cluster than HandWritten samples, resulting in higher accuracy when unaligned.
	\end{itemize}



 \begin{figure*}[t]
\centering
\begin{minipage}[t]{0.99\linewidth}
    \centering
    \subfloat{
        \includegraphics[width=1\linewidth]{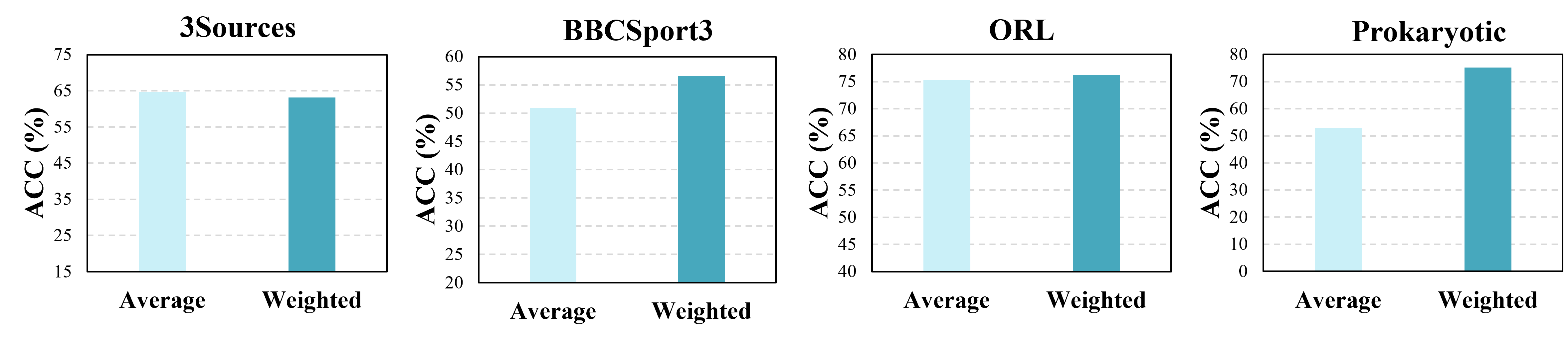}
    }
\end{minipage}
\qquad
\begin{minipage}[t]{0.99\linewidth}
    \centering
    \subfloat{
        \includegraphics[width=1\linewidth]{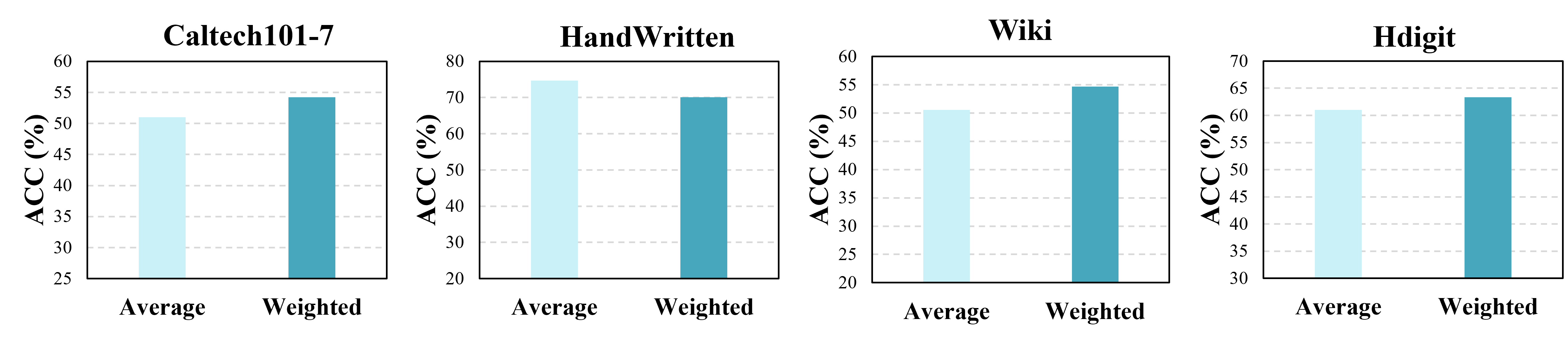}
    }
\end{minipage}

\caption{Performance comparison w./w.o. the view weight strategy of CoMSC on eight datasets. }
\label{comsc}
\end{figure*}

\begin{figure}[t]
    \centering
    \includegraphics[width =0.48\textwidth]{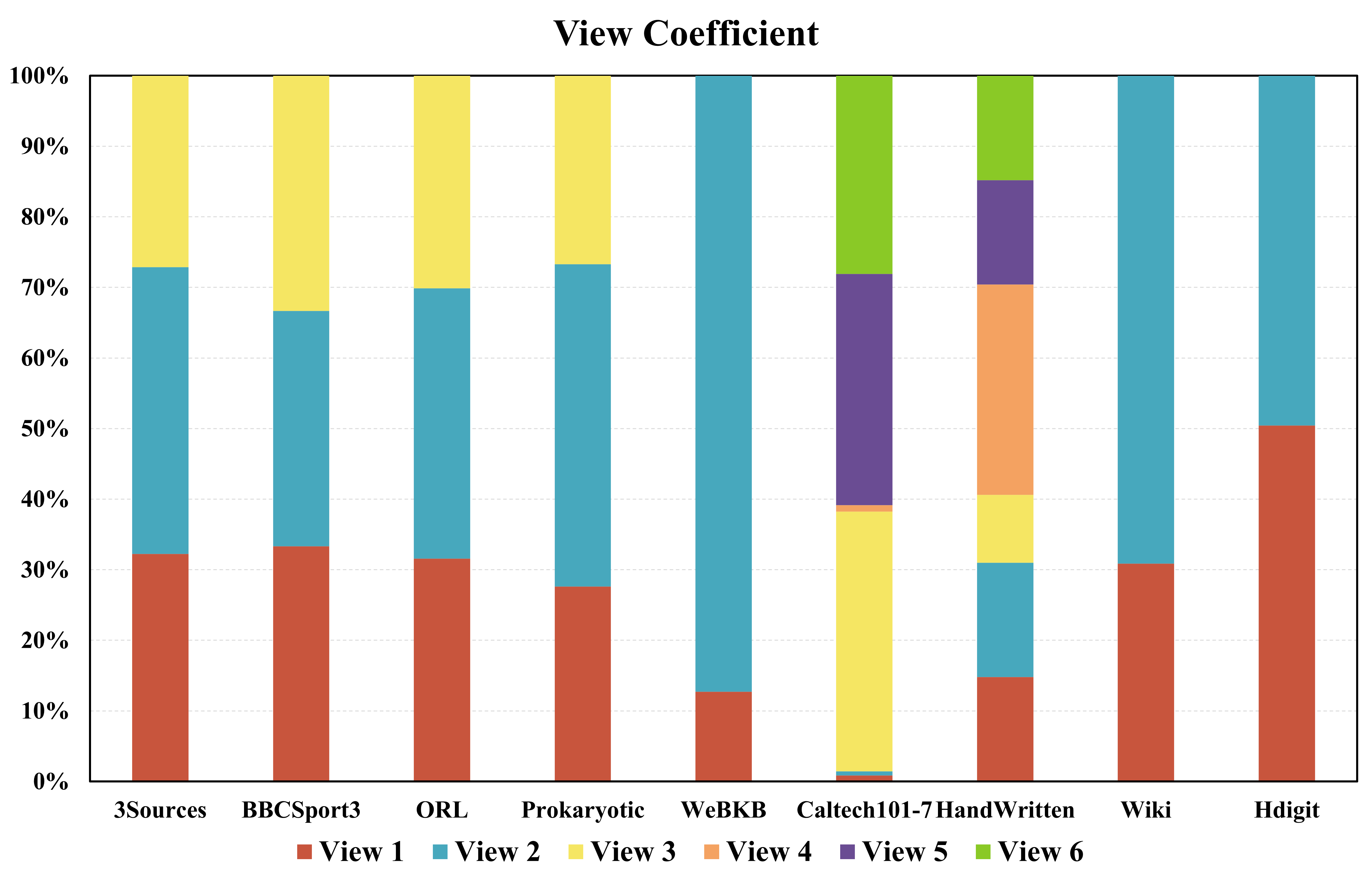}
    \caption{The square root of learned view coefficients on nine datasets with CoMSC}
    \label{fig:view_cofficients}
    
\end{figure}
\subsection{Experimental Results on Partially Paired Data}
As shown in Figure \ref{par_result}, for the purpose of investigating the performance of UPMGC-SM on partially paired data, we constructed data sets with different alignment proportions in each of the nine data sets, and the alignment proportions were increased from 10\% to 90\% with an interval of 10\%. Considering PVC and MvCLN can only handle the data with two views, we select the first two views of each dataset to adapt to our setting. To eliminate the effect of randomness, we randomly initialize each dataset 5 times and report the mean of the results. As depicted in Figure \ref{par_result}, we can observe that

\begin{itemize}
\item As the alignment ratio changes, the PVC results become unstable and might even show the worst performance under the fully paired circumstance, while the UPMGC-SM shows a more stable performance. We infer that this could be because neither PVC nor MvCLN provides a view selection strategy, the view quality of the first and second views is uncertain in our experimental setup, and the wrong alignment information learned in the poor view could mislead the final clustering partition.

	\item 
UPMGC-SM maintains superiority in each alignment ratio. Specifically, CoMSC+Ours achieved the best on most of the partially unpaired datasets, outperforming PVC by \textbf{16.18}\%, \textbf{24.63}\%, \textbf{27.80}\%, \textbf{4.63}\%, \textbf{30.08}\%,  \textbf{12.37}\%, \textbf{34.11}\%, \textbf{30.64}\%, \textbf{46.60}\% on the average of the nine datasets, respectively, which shows the superiority of UPMGC-SM on handling partially unpaired data.  we conjecture that it might be because it learns from the information of known paired data and fails to utilize the information of unpaired samples well, resulting in its poor performance on low-alignment rate datasets. 

	\item We noticed that in some datasets, PVC and MvCLN showed an increase in alignment ratio and a decrease in clustering performance instead. We conjecture that this may be caused by the fact that PVC and MvCLN can only select two views for alignment and cannot capture the complementary information of multiple views, which hinders the clustering performance.

 \item We observe that our method is insensitive to the alignment rate in some data with fewer views, such as WebKB and Wiki, and  we conjecture that on account of the quality differences among the views. Our model tends to learn information from better quality views, i.e., views that are selected, while poorer views are given a lower weight. In other words, the alignment of poorer views has less impact on our clustering results when the quality difference between views is considerable. The effectiveness of this view selection strategy can be verified in our later ablation experiments, as shown in Table \ref{selection}, where the clustering performance is significantly reduced when we select poorer views for alignment, which further illustrates the effectiveness of our selection principle for the alignment view.

	\end{itemize}
\subsection{Alignment Performance}
For the sake of evaluating the alignment accuracy of the proposed UPMGC-SM method, Similar to \cite{yu2021novel}, in this section, we choose the evaluation criterion MACC@10 to measure the accuracy of our cross-view alignment.
$$
\begin{aligned}
& \operatorname{MACC} @ q\left(\mathbf{P}^{mn}\right)=\frac{\sum_{i=1}^n \operatorname{count}\left(\mathbf{x}_i^{(m)}\right)}{n}, \\
& \operatorname{count}\left(\mathbf{x}_i^{(m)}\right)=\left\{\begin{array}{ll}
1, & \widehat{\mathbf{x}}_{(m, i)}^{(n)} \in C_{(m, i)}^n \\
0, & \text { others }
\end{array},\right.
\end{aligned}
$$
where $\mathbf{x}_i^{(m)}$ represents the $i$-th sample in the $m$-th view, $\widehat{\mathbf{x}}_{(m, i)}^n$ represents the real sample in the $n$-th view corresponding to the sample $\mathbf{x}_i^{(m)}$. $C_{(m, i)}^n$ consists
of the first q samples in the $n$-th view that $\mathbf{x}_i^{(m)}$ may be aligned according to $\mathbf{P}^{mn}$. In this paper, we use MACC@10 as a metric to evaluate the accuracy of our alignment. As is shown in Figure \ref{macc}, UPMGC-SM achieves good alignment accuracy on 3Sources and is marginally stable within 50 iterations on most datasets.  We obersve that the clustering results do not consistently show a positive correlation with the cross-view sample alignment rate, which can be illustrated from two aspects. First, a high alignment rate does not always lead to a better clustering performance. As the alignment rate increases, the number of unpaired samples becomes less. What is to say, we have less information available to learn the alignment matrix, which leads to poorer learning performance (because we only use unpaired samples in the alignment phase). PVC \cite{huang2020partially}, one of the prior works, shows a similar result with a high alignment ratio. Second, a lower alignment rate does not always lead to poorer clustering performance. Because clustering aims to enlarge the inter-cluster discrepancy and reduce the intra-cluster discrepancy, an over-refined alignment of the intra-cluster samples has little effect on the clustering performance, and sometimes a better clustering performance can be achieved with a rougher alignment (aligning to samples of the same cluster).

\subsection{Algorithm Convergence}
To evaluate the learning effectiveness, we record the variation objective value along with iterations. Due to space limitations, we only show the results on four datasets in Figure \ref{objection}. It can be observed that the objective function monotonically decreases and reaches convergence within 50 iterations.

\subsection{Selection Principle for the Alignment View}
  Here, we adopted the strategy of minimal reconstruction loss (RCM) to select the alignment view. We report the ablation experiments of our selection principle on nine benchmark datasets in Table \ref{selection}.
  From the results shown, our RCM strategy obtains better performance on the most dataset, demonstrating the effectiveness of the strategy. The reason is that the reconstruction loss can well describe the information of views by reflecting the gap between original feature data and reconstructed items. The view selection strategy is more evident on larger datasets, and we conjecture that the view quality is more important for finding a good alignment relationship due to the increased data.

\subsection{View Cofficients}
 Moreover, Figure  \ref{fig:view_cofficients} depicts the square root of the learned view coefficients $\alpha$ by the reconstruction loss. We observe that the distribution is non-sparse and exhibits discrepancy across multiple views, illustrating the effectiveness of our fusion strategy in integrating diverse and complementary information. Therefore, our UPMGC-SM can efficiently achieve graph fusion.

Figure \ref{comsc} shows the performance comparison w./w.o. the view weight strategy of CoMSC on eight datasets. "Average" indicates using the same weights for each view. From the experimental results, we can observe that the view weighting strategy can effectively improve the clustering performance in most datasets. In addition, comparable performance is achieved in other datasets, which illustrates the superiority of our view weighting strategy.

\section{Conclusion} 

In this paper, we propose a parameter-free alignment framework termed Unpaired Multi-View Graph Clustering with Cross-View Structure Matching (UPMGC-SM). UPMGC-SM adopts the fast projected fixed-point algorithm to effectively utilize the structure information of each view and establish the correct cross-view correspondences without any prior information. Moreover, we demonstrate the expandability and flexibility of the proposed method. However, the excessive time complexity restricts its application in large-scale scenarios. In the future, we will attempt to construct the bipartite graph approach to solve the DUP for large-scale scenarios effectively.

\ifCLASSOPTIONcompsoc
  \section*{Acknowledgments}
\else
  \section*{Acknowledgment}
\fi

This work was supported by the National Key R$\&$D Program of China 2020AAA0107100 and the National Natural Science Foundation of China (project no. 61872371, 61922088 and 61976196).

\ifCLASSOPTIONcaptionsoff
  \newpage
\fi



%

\bibliographystyle{IEEEtran}
\bibliography{myreference}

%

\begin{IEEEbiography}[{\includegraphics[width=1in,height=1.10in,clip,keepaspectratio]{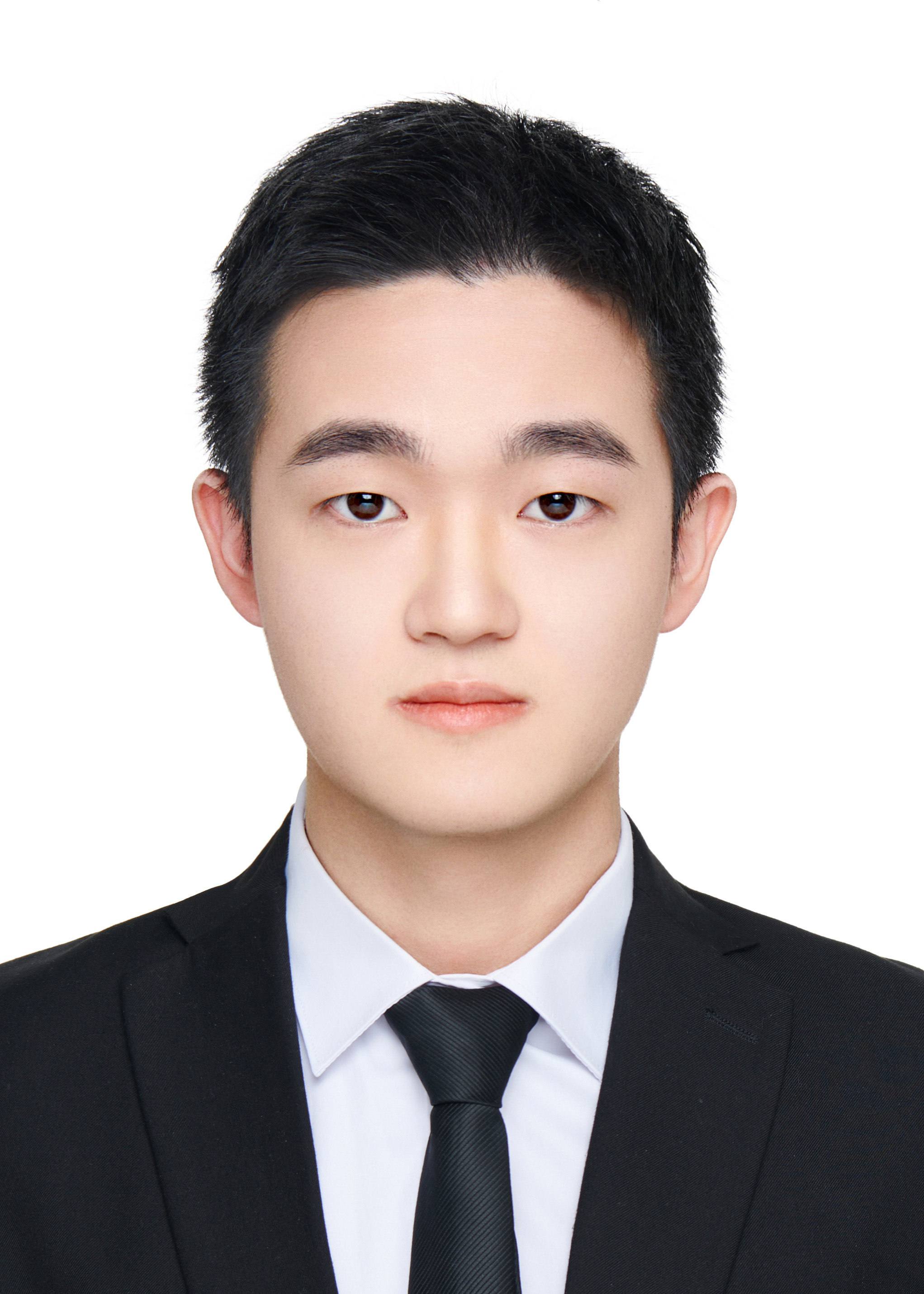}}]
{Yi Wen} is recommended for admission to the National University of Defense Technology (NUDT) as a master's student with  excellent grades and competition awards. He is working hard to pursue his master degree. His current research interests include multiple-view learning, scalable kernel k-means and graph representation learning.
\end{IEEEbiography}
\vspace{-5pt}

\begin{IEEEbiography}[{\includegraphics[width=1in,height=1.10in,clip,keepaspectratio]{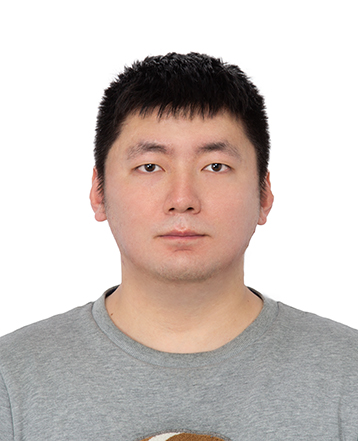}}]
{Siwei Wang} is currently pursuing the Ph.D. degree with the National University of Defense Technology (NUDT), China. He has published several papers and served as a PC Member/Reviewer
in top journals and conferences, such as IEEE
TRANSACTIONS ON KNOWLEDGE AND DATA ENGINEERING (TKDE), IEEE TRANSACTIONS ON NEURAL NETWORKS AND LEARNING SYSTEMS (TNNLS), IEEE TRANSACTIONS ON IMAGE PROCESSING (TIP), IEEE TRANSACTIONS ON
CYBERNETICS (TCYB), IEEE TRANSACTIONS ON
MULTIMEDIA (TMM), ICML, CVPR, ECCV, ICCV, AAAI, and IJCAI. His current research interests include kernel learning, unsupervised multiple-view
learning, scalable clustering, and deep unsupervised learning.
\end{IEEEbiography}
\vspace{-5pt}

\begin{IEEEbiography}[{\includegraphics[width=1in,height=1.05in,clip,keepaspectratio]{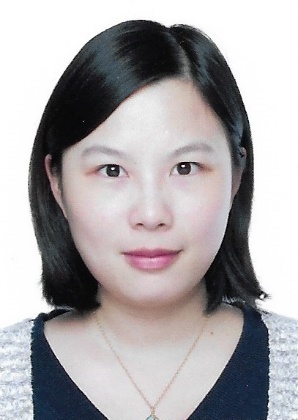}}]{Qing Liao} received her Ph.D. degree in computer science and engineering in 2016 supervised by Prof. Qian Zhang from the Department of Computer Science and Engineering of the Hong Kong University of Science and Technology. She is currently a professor with School of Computer Science and Technology, Harbin Institute of Technology (Shenzhen), China. Her research interests include artificial intelligence and data mining.
\end{IEEEbiography}

\begin{IEEEbiography}
[{\includegraphics[width=1in,height=1.15in,clip,keepaspectratio]{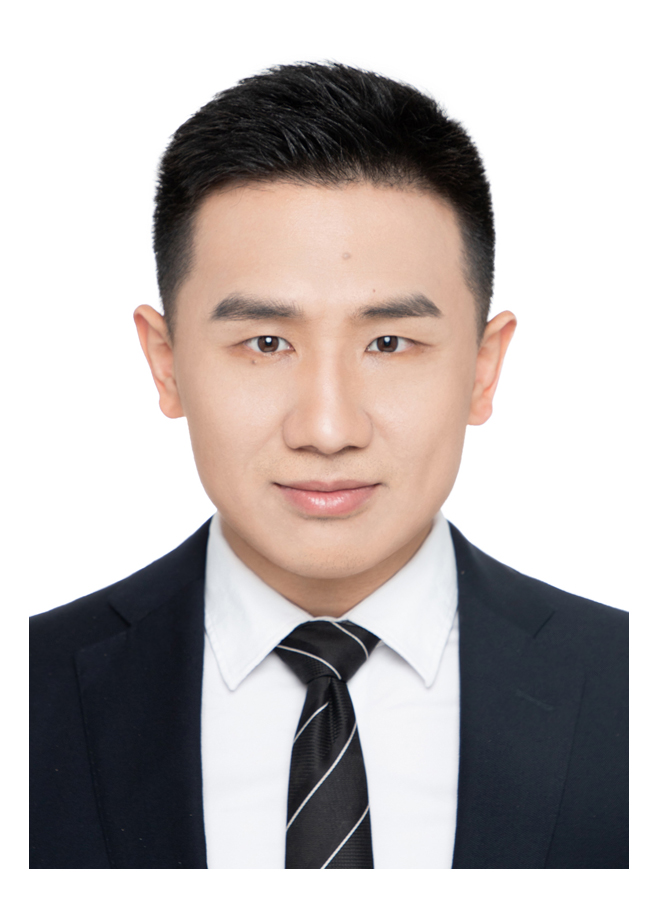}}]{Weixuan Liang} is pursuing his P.H.D degree in National University of Defense Technology (NUDT), China. He has authored or coauthored papers in journals and conferences, such as the IEEE Transactions on Knowledge and Data Engineering (TKDE), Annual Conference on Neural Information Processing Systems (NeurIPS), AAAI Conference on Artificial Intelligence (AAAI), and ACM Multimedia (ACM MM). His current research interests include multi-view clustering, learning theory, graph-based learning, and kernel methods.
\end{IEEEbiography}
\vspace{-10pt}

\begin{IEEEbiography}[{\includegraphics[width=1in,height=1.10in,clip,keepaspectratio]{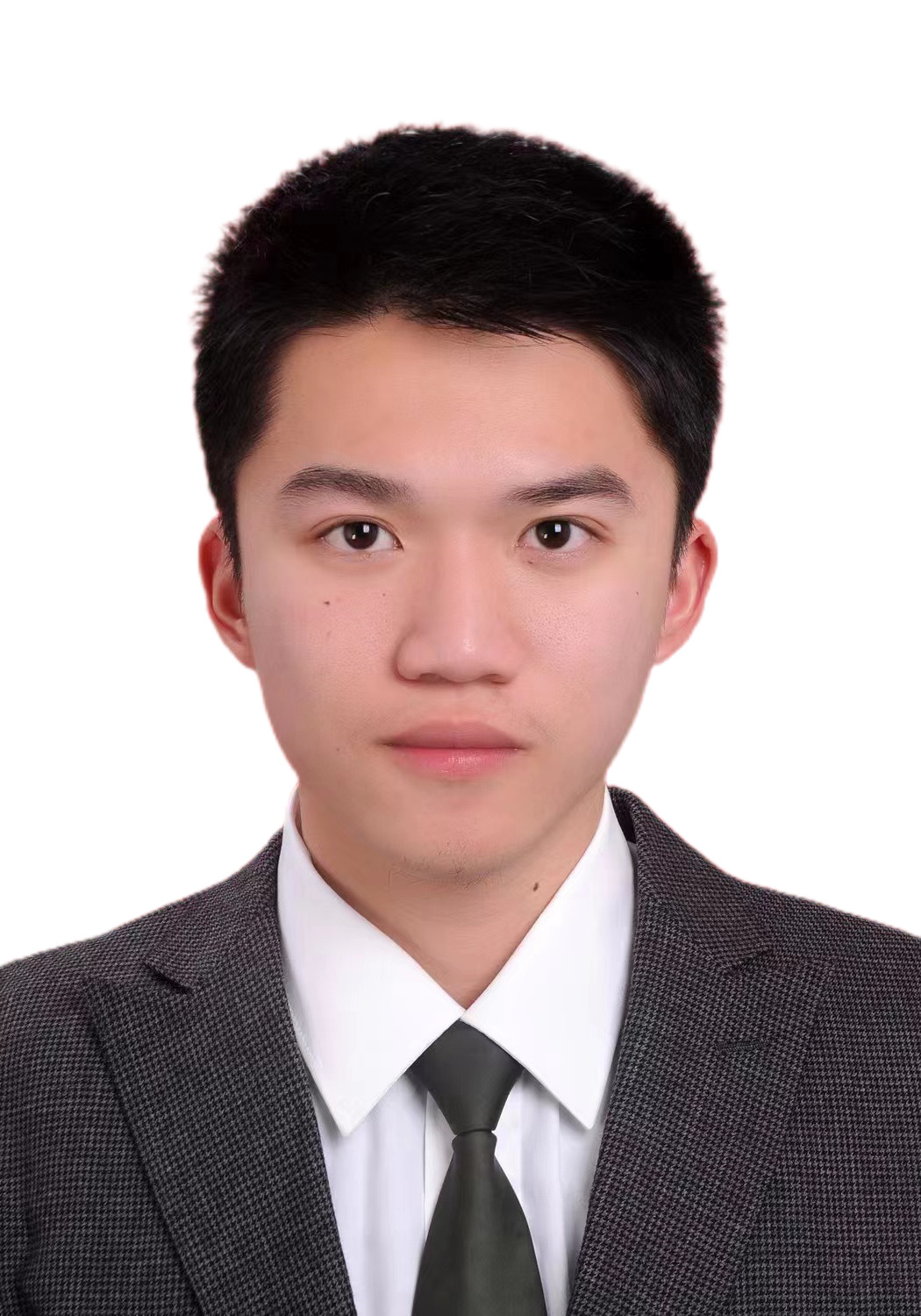}}]{Ke Liang} is currently pursuing a Ph.D. degree at the National University of Defense Technology (NUDT). Before joining NUDT, he got his BSc degree at Beihang University (BUAA) and received his MSC degree from the Pennsylvania State University (PSU). His current research interests include graph representation learning, multi-modal representation learning, and medical image processing.
\end{IEEEbiography}

\vspace{-10pt}
\begin{IEEEbiography}[{\includegraphics[width=1in,height=1.10in,clip,keepaspectratio]{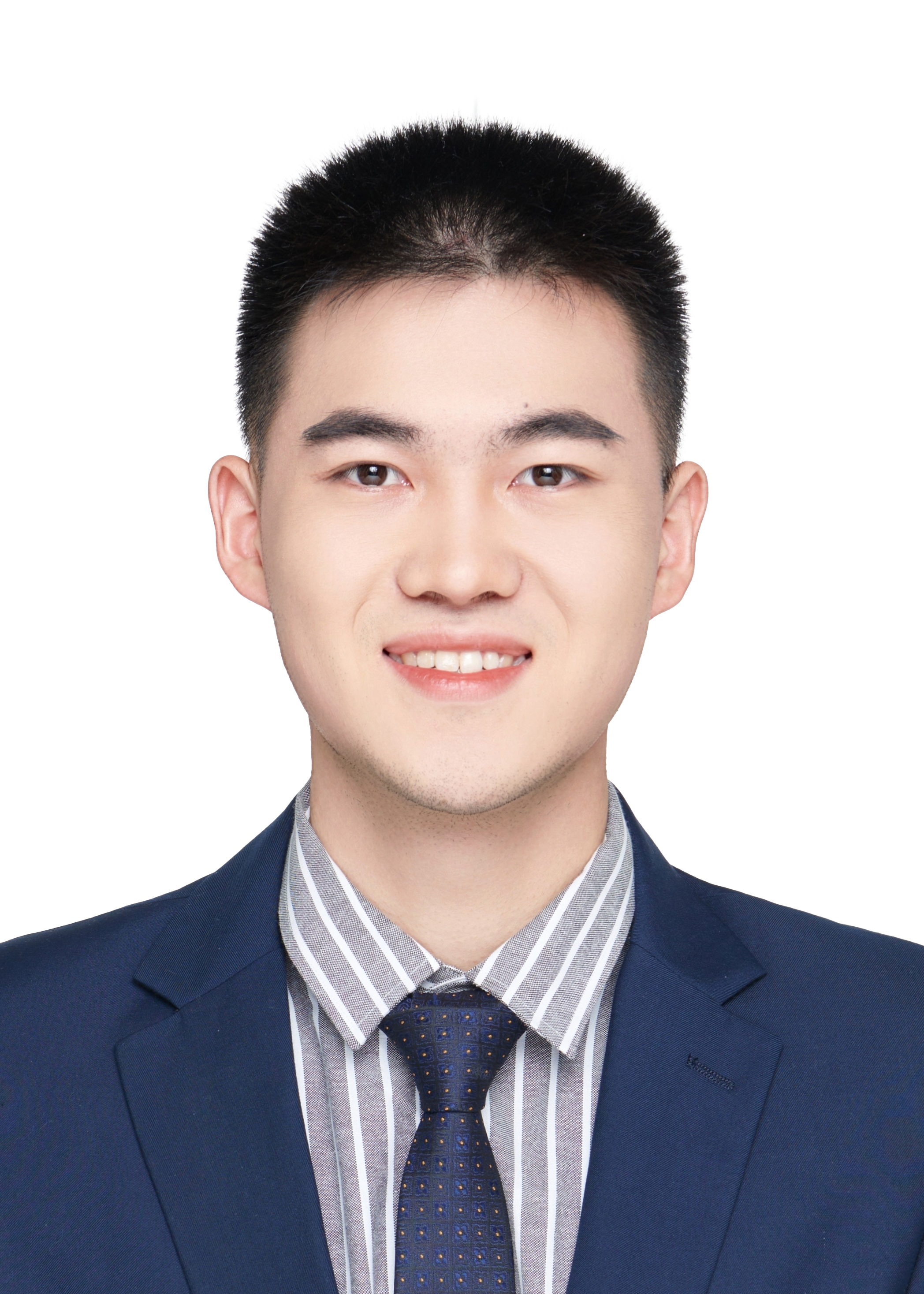}}]{Xinhang Wan}  received the B.E degree in Computer Science and Technology from Northeastern University, Shenyang, China, in 2021. He is currently working toward the Ph.D. degree with the National University of Defense Technology (NUDT), Changsha, China.
He has published papers in conferences, such as ACM MM and AAAI. His current research interests include multi-view learning, incomplete multi-view clustering, and continual clustering.
\end{IEEEbiography}

\begin{IEEEbiography}[{\includegraphics[width=1in,height=1.10in,clip,keepaspectratio]{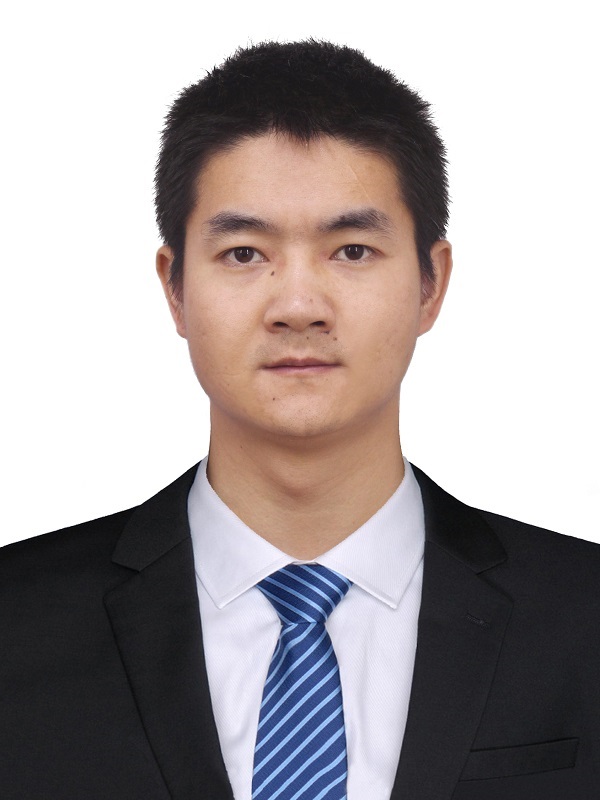}}]{Xinwang Liu} received his PhD degree from National University of Defense Technology (NUDT), China. He is now Professor of School of Computer, NUDT. His current research interests include kernel learning and unsupervised feature learning. Dr. Liu has published 60+ peer-reviewed papers, including those in highly regarded journals and conferences such as IEEE T-PAMI, IEEE T-KDE, IEEE T-IP, IEEE T-NNLS, IEEE T-MM, IEEE T-IFS, ICML, NeurIPS, ICCV, CVPR, AAAI, IJCAI, etc. He serves as the associated editor of Information Fusion Journal. More information can be found at \url{https://xinwangliu.github.io/}.
\end{IEEEbiography}





\end{document}